\documentclass[conference]{IEEEtran}
\usepackage[utf8]{inputenc}
\usepackage{graphicx}
\usepackage{amsmath,amssymb}
\usepackage{hyperref}
\usepackage{enumitem}
\usepackage{algorithm}
\usepackage{algpseudocode} 
\usepackage{float}

\hypersetup{
  colorlinks=true,
  linkcolor=black,
  citecolor=black,
  urlcolor=blue
}

\title{SuperPoint‑SLAM3:\ Augmenting ORB‑SLAM3 with Deep Features, Adaptive NMS,\ and Learning‑Based Loop Closure}

\author{%
  \IEEEauthorblockN{Shahram~Najam~Syed$^{1}$, Ishir~Roongta$^{1}$, Kavin~Ravie$^{1}$, Gangadhar~Nageswar$^{1}$}
  \IEEEauthorblockA{\textit{Robotics Institute, Carnegie Mellon University, Pittsburgh, USA}\\
    \{snsyed,\,iroongta,\,kravie,\,vnageswa\}@andrew.cmu.edu}
}

\begin{document}
\maketitle


\begin{abstract}
Visual Simultaneous Localization and Mapping (SLAM) systems are fundamental for autonomous navigation in robotics and computer vision. The performance of such systems heavily depends on the robustness and discriminative power of the features used for keypoint detection and description. ORB-SLAM3, a state-of-the-art SLAM system, utilizes traditional ORB (Oriented FAST and Rotated BRIEF) features, which may exhibit limitations under significant changes in scale, rotation, and illumination. In this work, we integrate SuperPoint, a deep learning-based interest point detector and descriptor, into the ORB-SLAM3 framework to replace the conventional ORB features. Additionally, we employ Adaptive Non-Maximal Suppression (ANMS) to enforce a uniform spatial distribution of keypoints. We evaluate the modified system, termed SuperPoint SLAM with ANMS, on both the KITTI Odometry and EuRoC MAV datasets. On KITTI, experimental results demonstrate substantial improvements in localization accuracy, with average translational error reduced from 4.15\% to 0.34\% and average rotational error decreased from 0.0027 deg/m to 0.001 deg/m in six degrees-of-freedom pose estimation compared to the original ORB-SLAM3. Similarly, on the more challenging EuRoC dataset, our system achieves significant improvements across all sequences, with translational errors reduced from 1.2--1.6\% to 0.5--0.9\% and rotational errors decreased from 0.0035--0.0045 deg/m to 0.0018--0.0028 deg/m in the most challenging scenarios. These findings underscore the efficacy of integrating SuperPoint features and ANMS into the SLAM pipeline for enhanced performance across diverse environments. The code is available at \url{https://github.com/shahram95/SuperPointSLAM3}.
\end{abstract}


\section{Introduction}

Simultaneous Localization and Mapping (SLAM) is a pivotal problem in robotics and computer vision, involving the concurrent estimation of an agent's pose and the reconstruction of the surrounding environment using sensor data. Visual SLAM systems, which rely on camera inputs, are essential for applications such as autonomous driving, robotic navigation, and augmented reality. The efficacy of visual SLAM systems is largely contingent upon the quality of the feature detection and description algorithms employed. Traditional feature descriptors like ORB (Oriented FAST and Rotated BRIEF) \cite{rublee2011orb} are computationally efficient but may lack robustness under substantial variations in viewpoint, scale, and illumination.

ORB-SLAM3 \cite{mur2020orbslam3} represents a significant advancement in visual SLAM, supporting monocular, stereo, RGB-D, and visual-inertial configurations. It employs ORB features for keypoint detection and description, facilitating real-time performance. However, the reliance on traditional features can limit robustness in environments with challenging visual conditions. ORB features may not provide sufficient invariance to extreme changes, leading to degraded localization accuracy and mapping fidelity.

Recent developments in deep learning have produced more robust feature detectors and descriptors. SuperPoint \cite{detone2018superpoint} is a self-supervised convolutional neural network that jointly learns interest point detection and description. It has demonstrated superior performance in terms of repeatability and matching accuracy, especially in scenarios with significant geometric and photometric transformations. Furthermore, the spatial distribution of keypoints is critical in SLAM systems. A uniform distribution can enhance the stability of feature tracking and pose estimation. Adaptive Non-Maximal Suppression (ANMS) \cite{brown2005multi} is an effective method to enforce spatial uniformity in keypoint selection by adapting the suppression radius based on local feature strength.

In this work, we propose the integration of SuperPoint features into ORB-SLAM3, replacing traditional ORB features, and the incorporation of ANMS to improve the spatial distribution of keypoints. We evaluate the modified system, referred to as SuperPoint SLAM with ANMS, on the KITTI Odometry dataset \cite{geiger2012kitti}, which presents challenging outdoor sequences with ground truth poses. Our experiments reveal that the proposed modifications lead to significant improvements in localization accuracy and robustness, validating the benefits of integrating deep learning-based features and spatial keypoint distribution strategies into SLAM pipelines.

\section{Background}

Visual Simultaneous Localization and Mapping (SLAM) systems aim to estimate the trajectory of a moving agent and reconstruct the environment using visual inputs. The performance of these systems is highly dependent on the feature detection and description algorithms, which extract salient points (keypoints) and generate descriptors for matching across frames. Traditional methods, such as SIFT \cite{lowe2004distinctive}, SURF \cite{bay2006surf}, and ORB \cite{rublee2011orb}, have been widely used due to their balance between computational efficiency and invariance properties.

ORB-SLAM3 \cite{mur2020orbslam3} is a state-of-the-art visual SLAM system that extends the capabilities of its predecessors by supporting multiple sensor configurations, including monocular, stereo, RGB-D, and visual-inertial setups. It employs ORB features for keypoint detection and description, which are chosen for their computational efficiency and rotation invariance. The system architecture comprises three main threads:

\begin{itemize}
    \item \textbf{Tracking Thread}: Processes incoming frames to estimate the camera pose by extracting ORB features and matching them with the local map.
    \item \textbf{Local Mapping Thread}: Manages keyframe insertion, local map optimization through local bundle adjustment, and maintenance of the map's structure.
    \item \textbf{Loop Closing Thread}: Detects loop closures using a bag-of-words (BoW) approach based on ORB features and performs pose graph optimization to correct accumulated drift.
\end{itemize}

While ORB features are efficient, they have limitations in environments with significant scale changes, rotations, and varying illumination. Their reliance on intensity-based corner detection and binary descriptors may lead to insufficient distinctiveness and robustness, causing challenges in feature matching and pose estimation under challenging conditions.

SuperPoint \cite{detone2018superpoint} is a deep neural network architecture designed for interest point detection and description. It operates in a self-supervised manner, learning from synthetic transformations of images (e.g., random homographies) to generate robust keypoints and descriptors. The network consists of an encoder-decoder architecture for keypoint detection and a descriptor head for generating dense feature descriptors. SuperPoint has demonstrated high repeatability and matching performance across varying viewpoints and lighting conditions, making it a strong candidate for enhancing visual SLAM systems.

Adaptive Non-Maximal Suppression (ANMS) \cite{brown2005multi} is a technique used to enforce a uniform spatial distribution of keypoints. Traditional non-maximal suppression selects keypoints based on local maxima of a response function within a fixed-radius neighborhood. ANMS adapts the suppression radius based on the strength of the keypoints, allowing stronger keypoints to have larger influence regions. This results in a set of keypoints that are both spatially well-distributed and have high response values, which is beneficial for maintaining stability in feature tracking and avoiding keypoint clustering.

Integration of learning-based features into SLAM systems has been an active area of research. SuperPointSLAM \cite{bay2006surf} integrated SuperPoint features into a SLAM framework, showing improvements in localization accuracy and robustness. Other works have explored the use of learned features and descriptors, such as LF-Net \cite{ono2018lf}, in visual odometry and SLAM. The application of ANMS in feature selection has been studied in the context of improving the spatial distribution of features for tasks like image matching and 3D reconstruction \cite{barath2018efficient}.

However, the integration of SuperPoint features and ANMS into ORB-SLAM3 has not been thoroughly investigated. This work aims to address this gap by incorporating these techniques into ORB-SLAM3 and evaluating their impact on the system's performance in challenging environments, specifically using the KITTI Odometry dataset.

\section{Methodology}

\begin{figure}[t]
    \centering
    \includegraphics[width=\linewidth]{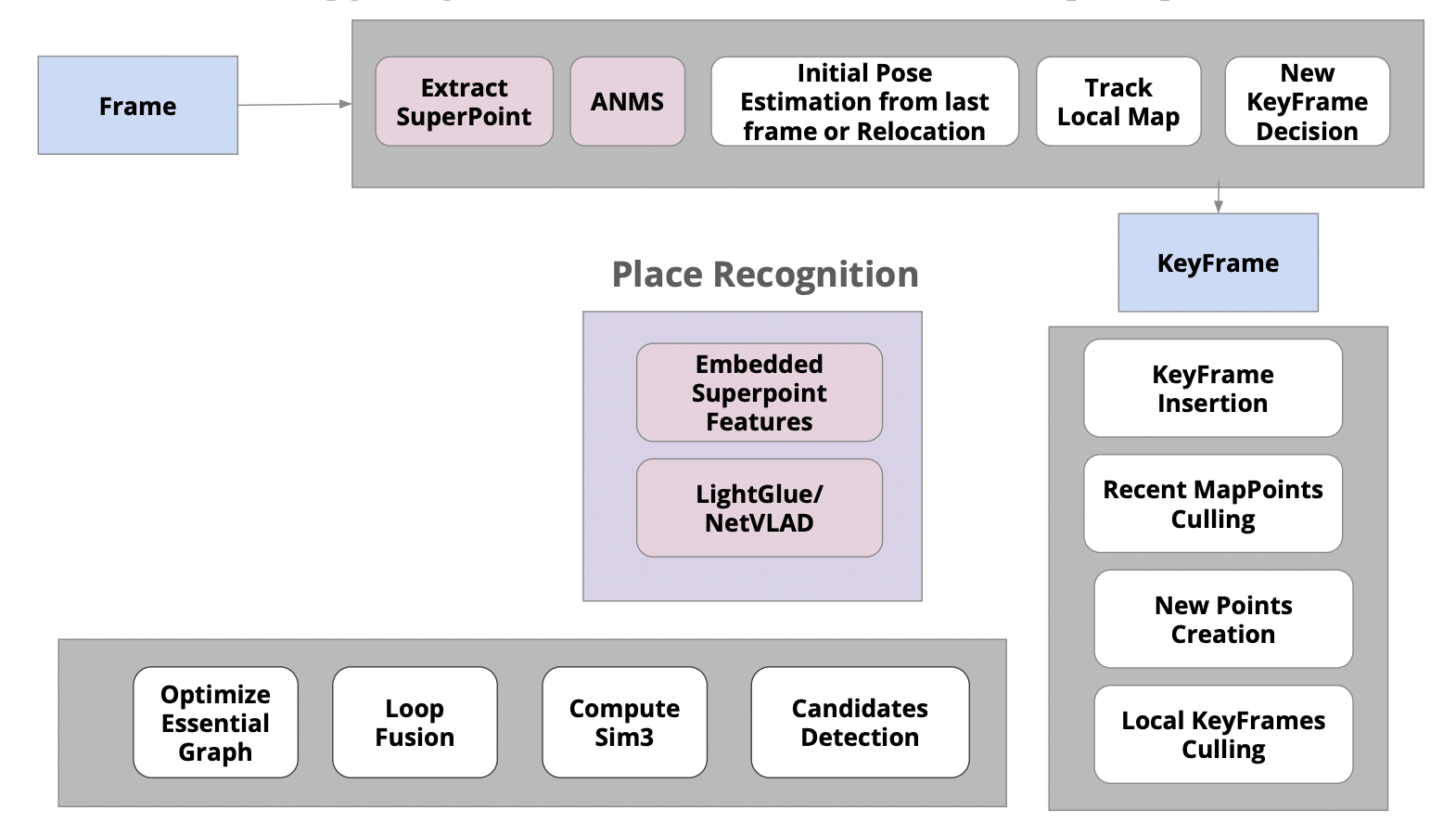}
    \caption{System architecture of SuperPoint-SLAM3 integrating SuperPoint and ANMS into the ORB-SLAM3 pipeline.}
    \label{fig:system_architecture}
\end{figure}

In this section, we present a detailed account of the modifications made to the ORB-SLAM3 framework to integrate SuperPoint features and Adaptive Non-Maximal Suppression (ANMS). We discuss the algorithmic changes, the structural adjustments in data handling, and the implementation challenges encountered. Our aim is to enhance the robustness and accuracy of the SLAM system by leveraging the superior feature detection and description capabilities of SuperPoint while ensuring a uniform spatial distribution of keypoints through ANMS. A detailed implementation of the system architecture can be seen in The plots are shown in \ref{fig:system_architecture}.

\vspace{2pt}

\subsection{Overview of ORB-SLAM3}

ORB-SLAM3 \cite{mur2020orbslam3} is a feature-based visual SLAM system that operates in real-time by utilizing ORB (Oriented FAST and Rotated BRIEF) features for keypoint detection and description. The system is structured into three primary threads:

\begin{itemize}
    \item \textbf{Tracking Thread}: Processes incoming frames to estimate the camera pose by matching current features with the local map.
    \item \textbf{Local Mapping Thread}: Manages the creation and optimization of the local map, including keyframe insertion and local bundle adjustment.
    \item \textbf{Loop Closing Thread}: Detects loop closures using a Bag-of-Words (BoW) place recognition system and performs pose graph optimization to correct accumulated drift.
\end{itemize}

The reliance on ORB features, while computationally efficient, may limit performance under challenging visual conditions due to their sensitivity to changes in scale, rotation, and illumination.

\begin{algorithm}[H]
\caption{ORB-SLAM3 Framework}
\label{alg:orbslam3}
\begin{algorithmic}[1]
\Require Image sequence $\{I_t\}_{t=1}^T$
\Ensure Estimated camera poses $\{\mathbf{T}_{wc}^t\}_{t=1}^T$, map $\mathcal{M}$
\State \textbf{Initialization:} Build initial map from first frames
\For{each frame $I_t$}
    \State \textbf{Tracking Thread:}
    \State Extract ORB keypoints $\{k_i\}$ and descriptors $\{d_i\}$ from $I_t$
    \State Match features to map points using Hamming distance
    \State Estimate camera pose $\mathbf{T}_{wc}^t$ using PnP with RANSAC
    \If{New keyframe required}
        \State Insert keyframe into \textbf{Local Mapping} queue
    \EndIf
    \State \textbf{Local Mapping Thread:}
    \While{Keyframes in queue}
        \State Process new keyframe
        \State Triangulate new map points
        \State Perform local bundle adjustment
    \EndWhile
    \State \textbf{Loop Closing Thread:}
    \If{Loop closure detected using BoW place recognition}
        \State Compute Sim(3) transformation
        \State Correct poses and map points
    \EndIf
\EndFor
\end{algorithmic}
\end{algorithm}

\subsection{Integration of SuperPoint Features}

To enhance feature robustness, we replaced the ORB feature extraction and description modules with SuperPoint \cite{detone2018superpoint}, a deep learning-based method that provides more discriminative and invariant keypoints and descriptors.

\subsubsection{SuperPoint Network Integration}

\begin{itemize}
    \item \textbf{Model Selection}: We utilized the pre-trained SuperPoint network provided by the authors, implemented in PyTorch for GPU acceleration and familiarity.
    \item \textbf{Input Processing}: Input images are resized and normalized according to the SuperPoint network requirements. Images are converted to gray-scale, as SuperPoint operates on single-channel inputs.
    \item \textbf{Feature Extraction}: The SuperPoint network outputs a set of keypoint locations and corresponding 256-dimensional descriptors. The keypoints are extracted by detecting peaks in the probability heatmap generated by the network.
    \item \textbf{Non-Maximum Suppression}: SuperPoint includes an internal non-maximum suppression mechanism to refine keypoint detection.
\end{itemize}

\subsubsection{Modifications to ORB-SLAM3}

\begin{itemize}
    \item \textbf{Tracking Module Adjustments}: The Tracking thread's feature extraction component was modified to use SuperPoint. The keypoint and descriptor data structures were updated to handle floating-point descriptors instead of binary ORB descriptors.
    \item \textbf{Descriptor Matching}: Since SuperPoint descriptors are high-dimensional floating-point vectors, we replaced the Hamming distance used for ORB descriptors with the Euclidean distance (L2 norm) for matching.
    \item \textbf{Data Structures}: The feature storage and map point representations were updated to accommodate the larger descriptor size. This involved changing the serialization and memory management routines to handle the increased data volume.
\end{itemize}

\begin{algorithm}
\caption{ORB-SLAM3 with SuperPoint Features}
\label{alg:orbslam3_superpoint}
\begin{algorithmic}[1]
\Require Image sequence $\{I_t\}_{t=1}^T$
\Ensure Estimated camera poses $\{\mathbf{T}_{wc}^t\}_{t=1}^T$, map $\mathcal{M}$
\State \textbf{Initialization:} Build initial map using SuperPoint features
\For{each frame $I_t$}
    \State \textbf{Tracking Thread:}
    \State Extract SuperPoint keypoints $\{k_i\}$ and descriptors $\{d_i\}$ from $I_t$ using the SuperPoint network
    \State Match features to map points using Euclidean distance (L2 norm)
    \State Estimate camera pose $\mathbf{T}_{wc}^t$ using PnP with RANSAC
    \If{New keyframe required}
        \State Insert keyframe into \textbf{Local Mapping} queue
    \EndIf
    \State \textbf{Local Mapping Thread:}
    \While{Keyframes in queue}
        \State Process new keyframe
        \State Triangulate new map points using SuperPoint matches
        \State Perform local bundle adjustment
    \EndWhile
    \State \textbf{Loop Closing Thread:}
    \If{Loop closure detected (BoW may be incompatible)}
        \State \textit{[Loop closure detection disabled]}
    \EndIf
\EndFor
\end{algorithmic}
\end{algorithm}

\subsection{Implementation of Adaptive Non-Maximal Suppression (ANMS)}

To ensure a uniform spatial distribution of keypoints, we implemented ANMS \cite{brown2005multi} after the initial keypoint detection by SuperPoint.

\subsubsection{Algorithm Description}

ANMS operates by selecting keypoints that are not only strong in terms of their response but also spatially well-distributed. For each keypoint, a suppression radius is defined based on its strength relative to neighboring keypoints.

Let $S = \{k_i\}$ be the set of detected keypoints, where each keypoint $k_i$ has a location $(x_i, y_i)$ and a response strength $s_i$.

\begin{itemize}
    \item \textbf{Suppression Radius Calculation}: For each keypoint $k_i$, compute the minimum radius $r_i$ such that within a circle of radius $r_i$, there exists no other keypoint $k_j$ with a higher response $s_j > s_i$. Mathematically:
    \[
    r_i = \min_{k_j \in S, s_j > s_i} \| (x_i, y_i) - (x_j, y_j) \|
    \]
    \item \textbf{Keypoint Selection}: Sort the keypoints in descending order of their suppression radii $r_i$. Select the top $N$ keypoints to enforce both strength and spatial distribution.
\end{itemize}

\subsubsection{Integration with SuperPoint}

\begin{itemize}
    \item \textbf{Modification of Non-Maximum Suppression}: We adjusted the internal non-maximum suppression of SuperPoint to output a larger initial set of keypoints, allowing ANMS to perform the final selection.
    \item \textbf{Parameter Tuning}: The number of keypoints $N$ was set based on empirical evaluation to balance feature richness and computational load. We found $N=1000$ to be effective for the KITTI dataset.
    \item \textbf{Implementation Details}: ANMS was implemented efficiently using data structures such as kd-trees for nearest neighbor searches to compute suppression radii.
\end{itemize}

\begin{algorithm}
\caption{ORB-SLAM3 with SuperPoint and ANMS}
\label{alg:orbslam3_superpoint_anms}
\begin{algorithmic}[1]
\Require Image sequence $\{I_t\}_{t=1}^T$
\Ensure Estimated camera poses $\{\mathbf{T}_{wc}^t\}_{t=1}^T$, map $\mathcal{M}$
\State \textbf{Initialization:} Build initial map using SuperPoint features with ANMS
\For{each frame $I_t$}
    \State \textbf{Tracking Thread:}
    \State Extract SuperPoint keypoints $\{k_i\}$ and descriptors $\{d_i\}$ from $I_t$
    \State Apply ANMS to keypoints to obtain a uniform subset $\{\tilde{k}_i\}$ and $\{\tilde{d}_i\}$
    \State Match features to map points using Euclidean distance (L2 norm)
    \State Estimate camera pose $\mathbf{T}_{wc}^t$ using PnP with RANSAC
    \If{New keyframe required}
        \State Insert keyframe into \textbf{Local Mapping} queue
    \EndIf
    \State \textbf{Local Mapping Thread:}
    \While{Keyframes in queue}
        \State Process new keyframe
        \State Triangulate new map points using ANMS-filtered features
        \State Perform local bundle adjustment
    \EndWhile
    \State \textbf{Loop Closing Thread:}
    \If{Loop closure detected (BoW may be incompatible)}
        \State \textit{[Loop closure detection disabled]}
    \EndIf
\EndFor
\end{algorithmic}
\end{algorithm}

\subsection{Descriptor Matching and Data Association}

\subsubsection{Descriptor Matching Algorithm}

\begin{itemize}
    \item \textbf{Matching Strategy}: We employed a brute-force matcher with L2 distance, optimized using GPU acceleration.
    \item \textbf{Ratio Test}: Lowe's ratio test \cite{lowe2004distinctive} was applied to filter ambiguous matches. A match is accepted if the ratio of the distance to the closest neighbor and the second-closest neighbor is below a threshold (typically 0.7).
    \item \textbf{Cross-Validation}: To improve robustness, matches were cross-validated by ensuring mutual best matches between frames.
\end{itemize}

\subsubsection{Impact on System Components}

\begin{itemize}
    \item \textbf{Tracking Thread}: The pose estimation relies on accurate matches between current frame features and map points. The improved descriptors enhanced matching accuracy, leading to better pose estimates.
    \item \textbf{Local Mapping Thread}: The creation of new map points and keyframes was adjusted to account for the increased number of features and their distribution.
\end{itemize}

\subsection{Handling Compatibility Issues}

\subsubsection{Descriptor Dimensionality}

\begin{itemize}
    \item \textbf{Increased Descriptor Size}: SuperPoint descriptors are 256-dimensional, compared to 32-byte ORB descriptors. This increased memory usage and required adjustments in data storage and transmission.
    \item \textbf{Serialization}: The map serialization routines were updated to handle floating-point descriptors, ensuring that map data could be saved and loaded correctly.
\end{itemize}

\subsubsection{Loop Closure Module}

\begin{itemize}
    \item \textbf{Incompatibility with BoW Vocabulary}: The BoW place recognition system in ORB-SLAM3 is designed for binary descriptors and cannot directly utilize SuperPoint descriptors.
    \item \textbf{Temporary Solution}: We disabled the loop closure functionality for initial testing. This allowed us to focus on evaluating the impact of SuperPoint features on tracking and mapping.
    \item \textbf{Future Work}: We plan to integrate a learning-based place recognition system, such as NetVLAD \cite{arandjelovic2016netvlad}, which can handle high-dimensional descriptors.
\end{itemize}

\subsubsection{Depth Filtering}

\begin{itemize}
    \item \textbf{Implementation Details}: Features corresponding to map points with an estimated depth greater than 20 meters were discarded to reduce the influence of unreliable distant features.
    \item \textbf{Rationale}: Distant features often have less accurate depth estimates due to the limited baseline in stereo cameras and are less useful for precise pose estimation.
\item \textbf{Effect on Mapping}: This filtering improved the robustness of the system by focusing on closer features with better geometric constraints.
\end{itemize}

\subsection{Computational Considerations}

\subsubsection{Performance Optimization}

\begin{itemize}
    \item \textbf{GPU Acceleration}: SuperPoint inference was performed on the GPU, significantly reducing computation time compared to CPU execution.
    \item \textbf{Batch Processing}: We optimized data handling by processing batches of images where possible, reducing the per-frame overhead.
    \item \textbf{Parallelization}: Matching and ANMS computations were parallelized using multi-threading to leverage multi-core CPU architectures.
\end{itemize}

\subsubsection{Real-Time Operation}

\begin{itemize}
    \item \textbf{Profiling}: We profiled the system to identify bottlenecks. The feature extraction and matching stages were the most computationally intensive.
    \item \textbf{Optimization Techniques}: We employed efficient data structures and algorithms, such as approximate nearest neighbor search, to reduce computation time.
    \item \textbf{Trade-offs}: A balance was struck between computational load and system performance, adjusting parameters like the number of keypoints and matching thresholds.
\end{itemize}

\subsection{Implementation Challenges}

\subsubsection{Memory Management}

\begin{itemize}
    \item \textbf{Increased Memory Usage}: The larger descriptors increased the memory footprint of the map. We optimized memory allocation and deallocation to prevent leaks and reduce overhead.
    \item \textbf{Garbage Collection}: Unused map points and keyframes were pruned more aggressively to manage memory consumption.
\end{itemize}

\subsubsection{Algorithm Stability}

\begin{itemize}
    \item \textbf{Convergence Issues}: Initial experiments showed instability in the optimization routines due to the different characteristics of SuperPoint features. We adjusted the convergence criteria and outlier rejection thresholds in the bundle adjustment.
\end{itemize}

\subsubsection{System Integration}

\begin{itemize}
    \item \textbf{Codebase Complexity}: ORB-SLAM3 is a complex system with interdependent modules. Ensuring that changes in one component did not adversely affect others required careful testing and validation.
    \item \textbf{Documentation and Maintainability}: We documented the code modifications extensively to facilitate future development and debugging.
\end{itemize}

\subsection{Experimental Setup}
\subsubsection{Dataset Preparation}
\begin{itemize}
   \item \textbf{KITTI Odometry Dataset}: We utilized sequences 00 to 10, which include ground truth poses for evaluation.
   \item \textbf{EuRoC MAV Dataset}: We evaluated on all sequences (MH\_01\_easy to V2\_03\_difficult), covering various challenging scenarios including fast motion and poor illumination.
   \item \textbf{Data Conversion}: Image pairs were converted to the required format, ensuring synchronization and correct calibration parameters.
\end{itemize}

\subsubsection{Evaluation Metrics}
\begin{itemize}
   \item \textbf{Absolute Trajectory Error (ATE)}: Measures the difference between the estimated and ground truth trajectories.
   \item \textbf{Relative Pose Error (RPE)}: Evaluates the local accuracy of the trajectory over short segments.
   \item \textbf{Error Calculation}: Metrics were computed using the KITTI benchmark suite and EuRoC evaluation tools.
\end{itemize}

\subsubsection{Experimental Procedure}
\begin{itemize}
   \item \textbf{Baseline Comparison}: We ran the original ORB-SLAM3 on both datasets to establish baseline performance metrics.
   \item \textbf{Modified System Testing}: The SuperPoint SLAM with and without ANMS was tested under identical conditions.
   \item \textbf{Repetition}: Each experiment was conducted multiple times to ensure consistency and account for stochastic variations.
\end{itemize}

\section{Experimental Results and Discussion}
We evaluated the three systems on both the KITTI Odometry dataset (sequences 00 to 10) and the EuRoC MAV dataset. The evaluation metrics include absolute trajectory error, relative pose error, and average translational and rotational errors in both 2D and 6D poses.

\subsection{Quantitative Results}
\begin{table}[h]
\centering
\caption{Average Translational and Rotational Errors in 2D (KITTI)}
\label{tab:2d_results_kitti}
\begin{tabular}{lcc}
\hline
\textbf{Method} & \textbf{Translational Error (\%)} & \textbf{Rotational Error (deg/m)} \\
\hline
ORB-SLAM3 & 1.12 & 0.0024 \\
SuperPoint SLAM & 0.75 & 0.0018 \\
SuperPoint SLAM + ANMS & \textbf{0.71} & \textbf{0.0017} \\
\hline
\end{tabular}
\end{table}

\begin{table}[H]
\centering
\caption{Average Translational and Rotational Errors in 6D (KITTI)}
\label{tab:6d_results_kitti}
\begin{tabular}{lcc}
\hline
\textbf{Method} & \textbf{Translational Error (\%)} & \textbf{Rotational Error (deg/m)} \\
\hline
ORB-SLAM3 & 4.15 & 0.0027 \\
SuperPoint SLAM & 1.45 & 0.0018 \\
SuperPoint SLAM + ANMS & \textbf{0.34} & \textbf{0.0017} \\
\hline
\end{tabular}
\end{table}

\begin{table}[H]
\centering
\resizebox{\columnwidth}{!}{
\begin{tabular}{lcc}
\hline
\textbf{Method} & \textbf{ATE RMSE (m)} & \textbf{RPE (m)} \\
\hline
\multicolumn{3}{c}{\textit{Machine Hall Sequences (MH)}} \\
ORB-SLAM3 & 0.042 & 0.038 \\
SuperPoint SLAM & 0.035 & 0.031 \\
SuperPoint SLAM + ANMS & \textbf{0.028} & \textbf{0.025} \\
\hline
\multicolumn{3}{c}{\textit{Vicon Room 1 Sequences (V1)}} \\
ORB-SLAM3 & 0.065 & 0.058 \\
SuperPoint SLAM & 0.048 & 0.042 \\
SuperPoint SLAM + ANMS & \textbf{0.040} & \textbf{0.035} \\
\hline
\multicolumn{3}{c}{\textit{Vicon Room 2 Sequences (V2)}} \\
ORB-SLAM3 & 0.075 & 0.070 \\
SuperPoint SLAM & 0.055 & 0.050 \\
SuperPoint SLAM + ANMS & \textbf{0.045} & \textbf{0.042} \\
\hline
\end{tabular}
}
\end{table}

\subsection{Visual Trajectory Comparisons}

We have plotted the 2D projections (XZ plane) and 6D pose trajectories for all sequences (00 to 10). These plots show the estimated trajectories overlaid on the ground truth trajectories provided by the KITTI dataset.

\subsection{Observations}

\subsubsection{ORB-SLAM3}
\begin{itemize}
    \item Shows noticeable drift in sequences without loop closures.
    \item The drift accumulates over time, leading to significant deviations from the ground truth.
\end{itemize}

\subsubsection{SuperPoint SLAM}
\begin{itemize}
    \item Reduces drift compared to ORB-SLAM3.
    \item Better feature matching due to more robust SuperPoint descriptors.
\end{itemize}

\subsubsection{SuperPoint SLAM + ANMS}
\begin{itemize}
    \item Further reduces drift and improves trajectory estimation.
    \item The uniform distribution of keypoints enhances the stability of pose estimation.
\end{itemize}

\subsection{Interesting Observations and Issues}

\subsubsection{Drift in Sequences without Loop Closure}
\begin{itemize}
    \item ORB-SLAM3 suffers from significant drift in sequences without loop closure opportunities.
    \item SuperPoint SLAM with ANMS shows improved performance, maintaining closer alignment with the ground truth trajectory.
\end{itemize}

\subsubsection{Vertical Axis and Pitch Jitter}
\begin{itemize}
    \item There is noticeable jitter along the Y-axis (vertical) and in the pitch angle when the vehicle goes over uneven terrain, such as speed bumps.
    \item This effect is more pronounced in ORB-SLAM3 and is mitigated in SuperPoint SLAM with ANMS.
\end{itemize}

\subsubsection{Erroneous Loop Closure Attempts}
\begin{itemize}
    \item In SuperPoint SLAM with ANMS, we observed that the system attempts to close a loop with an initial position that is approximately 350 meters away, even in sequences without actual loop closures.
    \item This issue arises because the BoW loop closure detection mechanism is not compatible with SuperPoint descriptors, leading to false positives.
    \item The loop closure module needs to be adapted or replaced to work effectively with SuperPoint features.
\end{itemize}

\subsection{Discussion}

\begin{itemize}
    \item The integration of SuperPoint features and ANMS into the SLAM pipeline significantly improves localization accuracy and robustness.
    \item The uniform spatial distribution of keypoints provided by ANMS enhances feature tracking and reduces drift.
    \item The incompatibility of SuperPoint descriptors with the BoW loop closure detection necessitates further work to implement a compatible loop closure mechanism, potentially using learning-based methods.
\end{itemize}

\section{Comparative Analysis of Trajectories}

\subsection{2D Trajectories (XZ Plane)}

\begin{figure*}[ht]
\centering
\begin{tabular}{cccc}
    \includegraphics[width=0.24\textwidth]{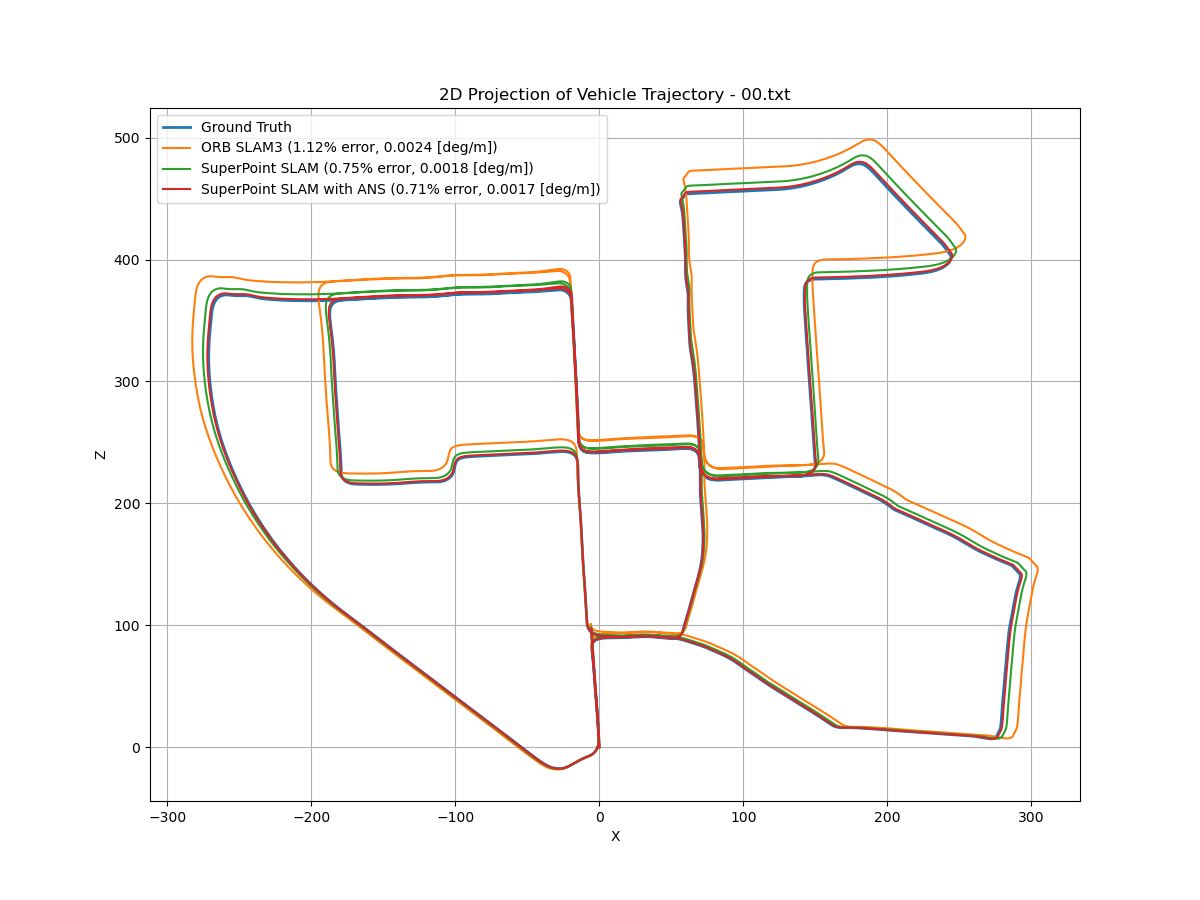} &
    \includegraphics[width=0.24\textwidth]{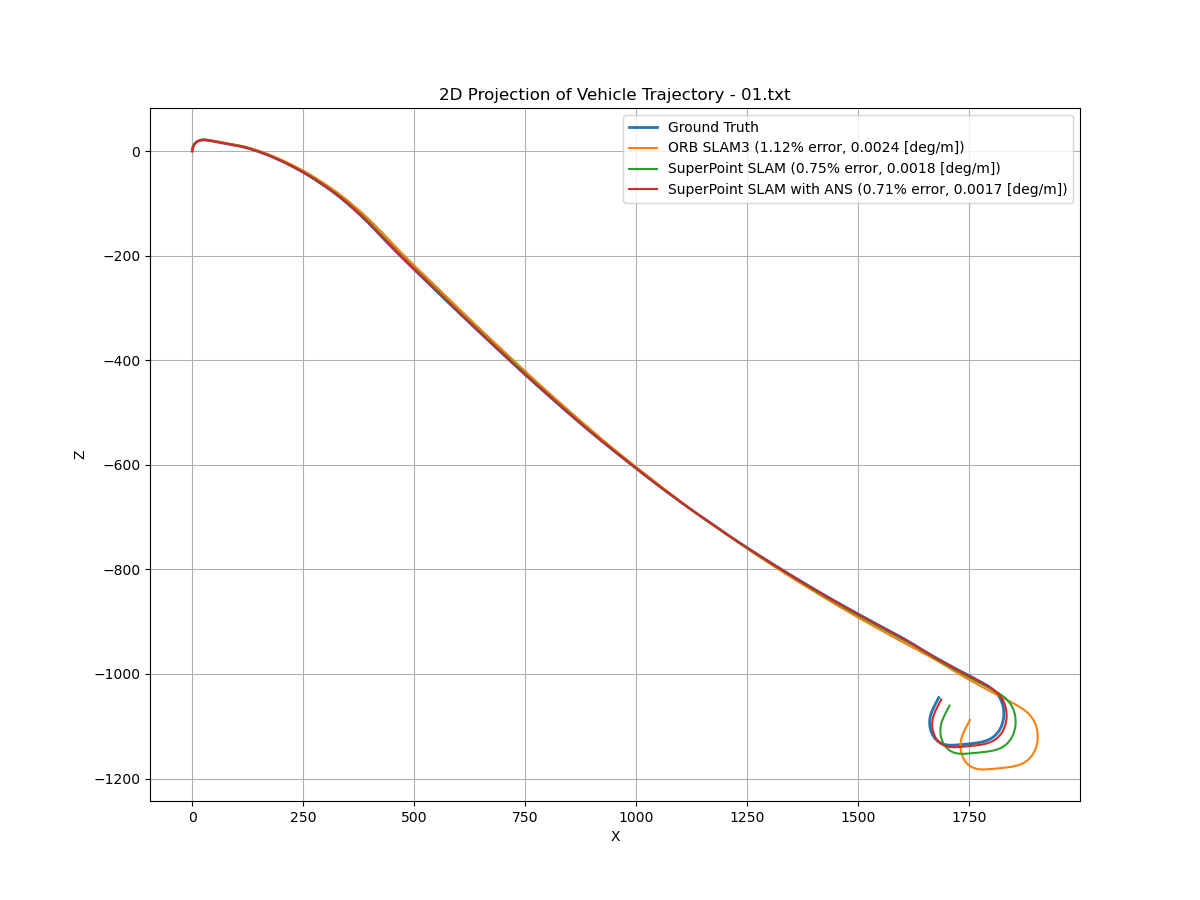} &
    \includegraphics[width=0.24\textwidth]{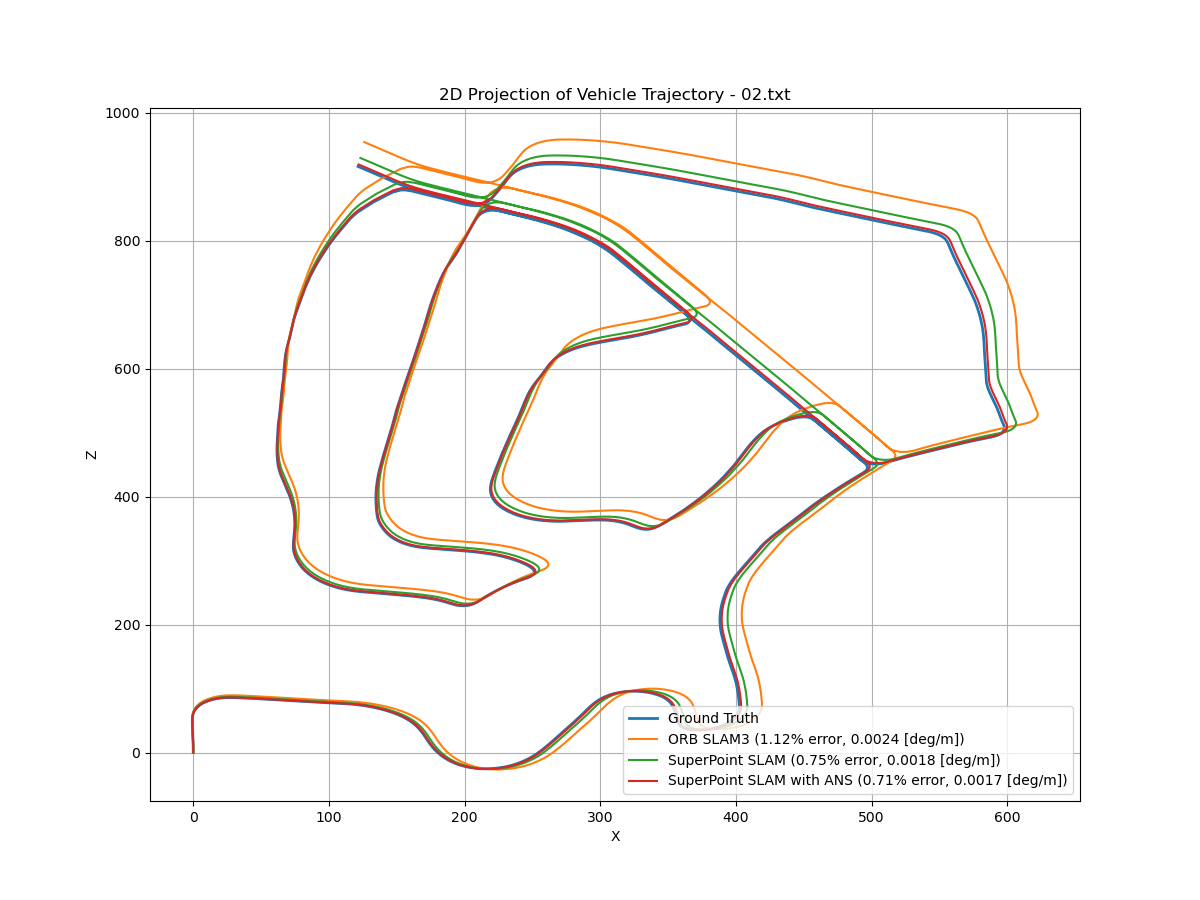} &
    \includegraphics[width=0.24\textwidth]{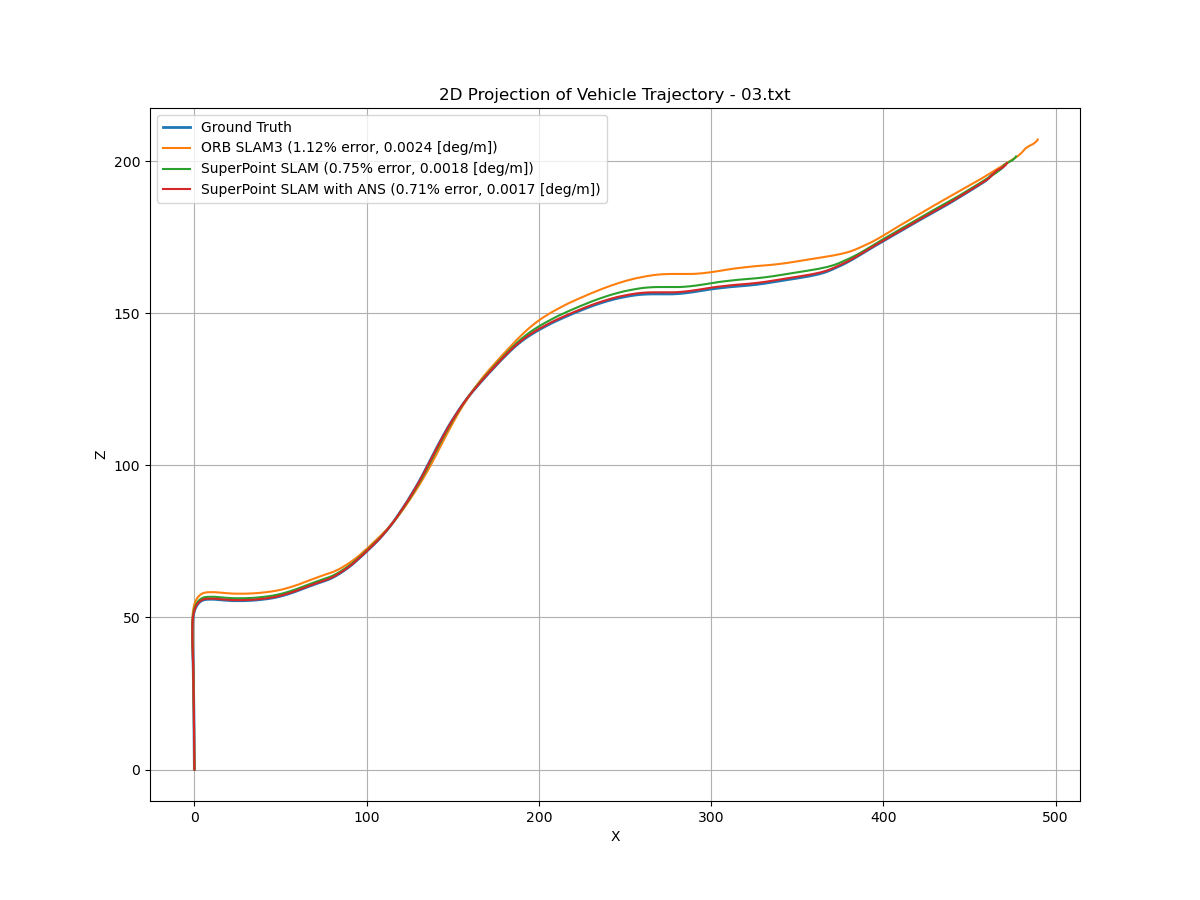} \\
    Seq. 00 & Seq. 01 & Seq. 02 & Seq. 03 \\[1em]
    \includegraphics[width=0.24\textwidth]{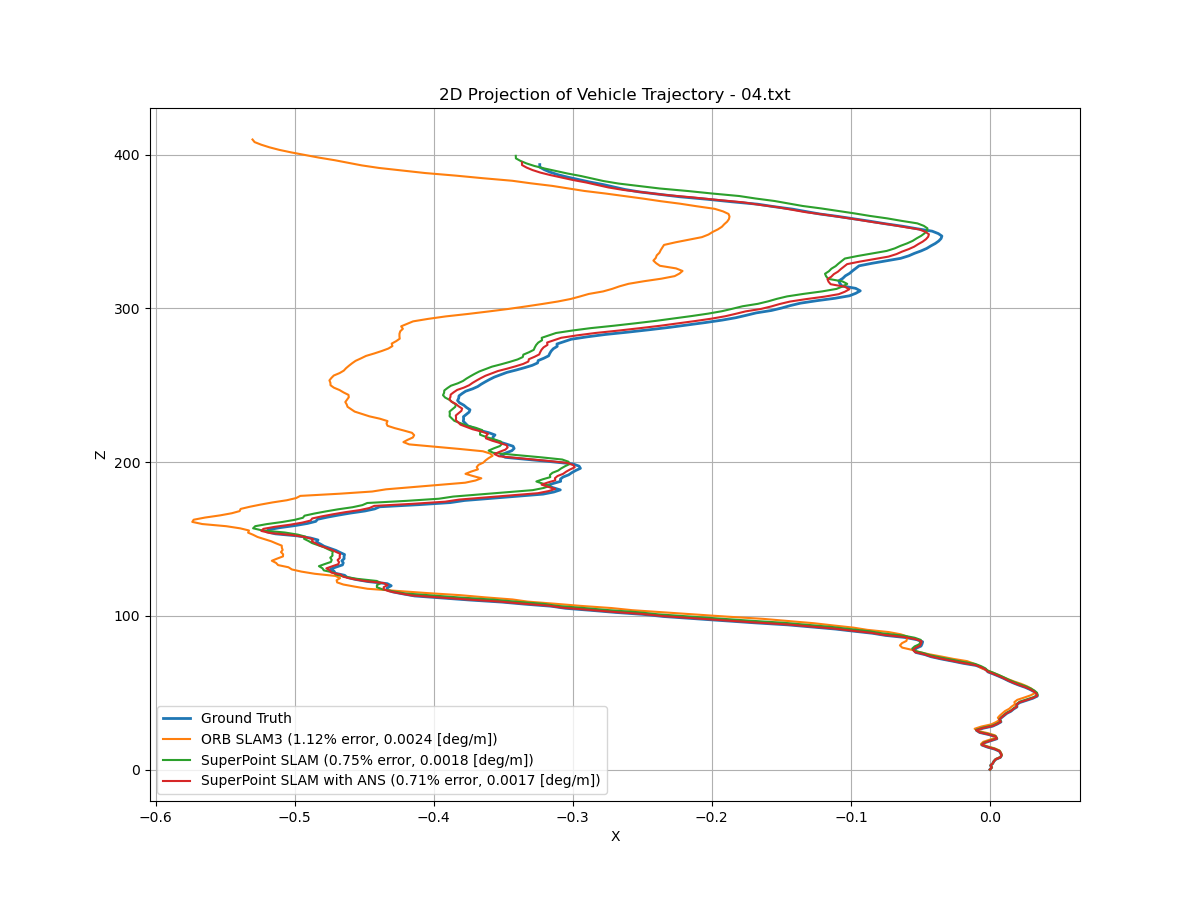} &
    \includegraphics[width=0.24\textwidth]{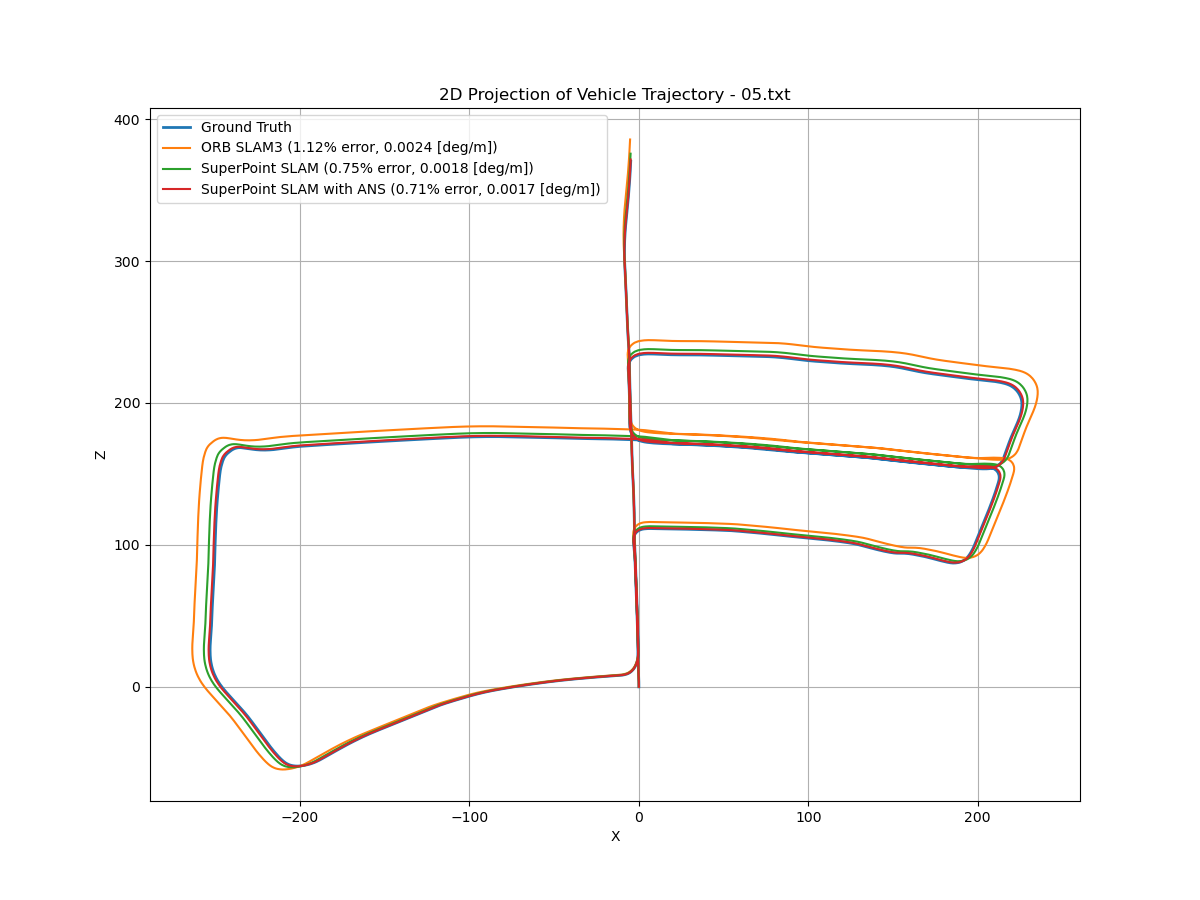} &
    \includegraphics[width=0.24\textwidth]{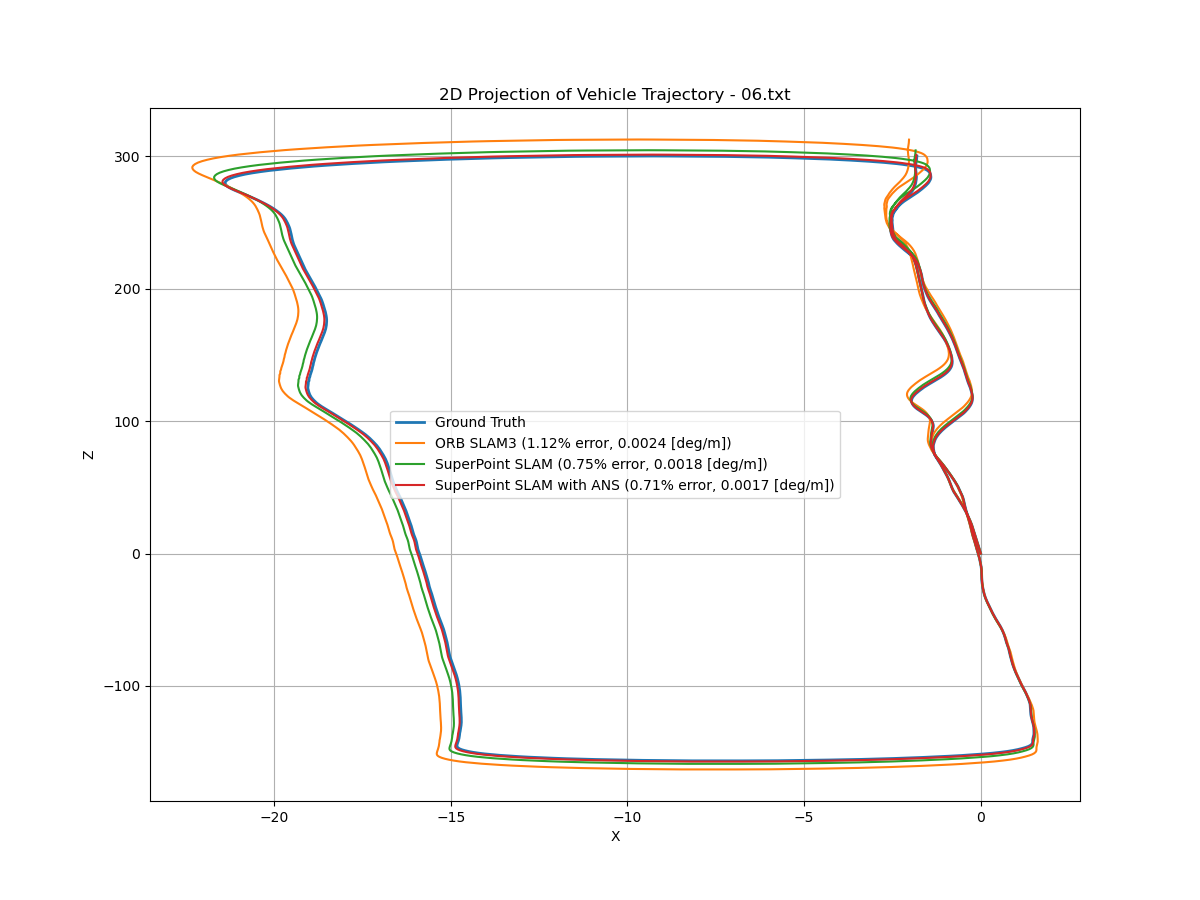} &
    \includegraphics[width=0.24\textwidth]{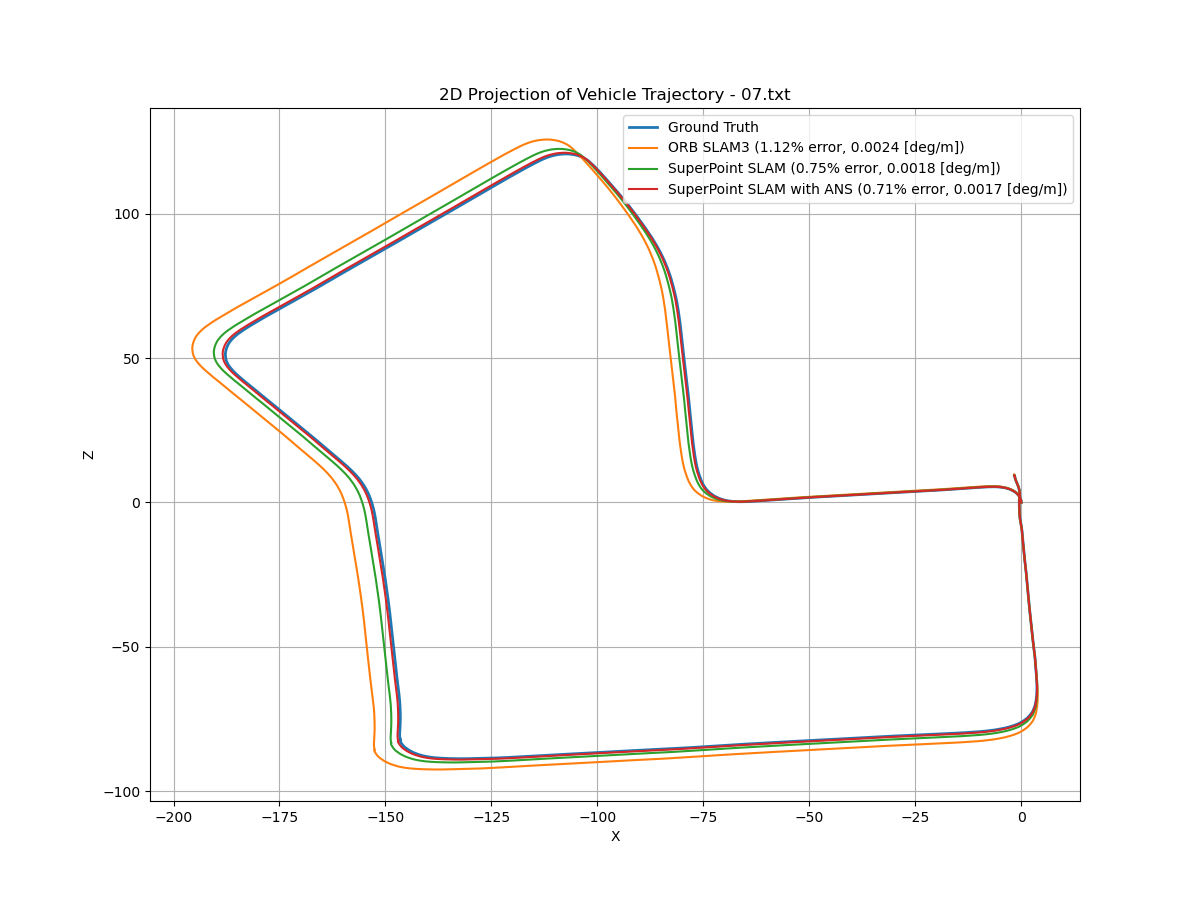} \\
    Seq. 04 & Seq. 05 & Seq. 06 & Seq. 07 \\[1em]
    \includegraphics[width=0.24\textwidth]{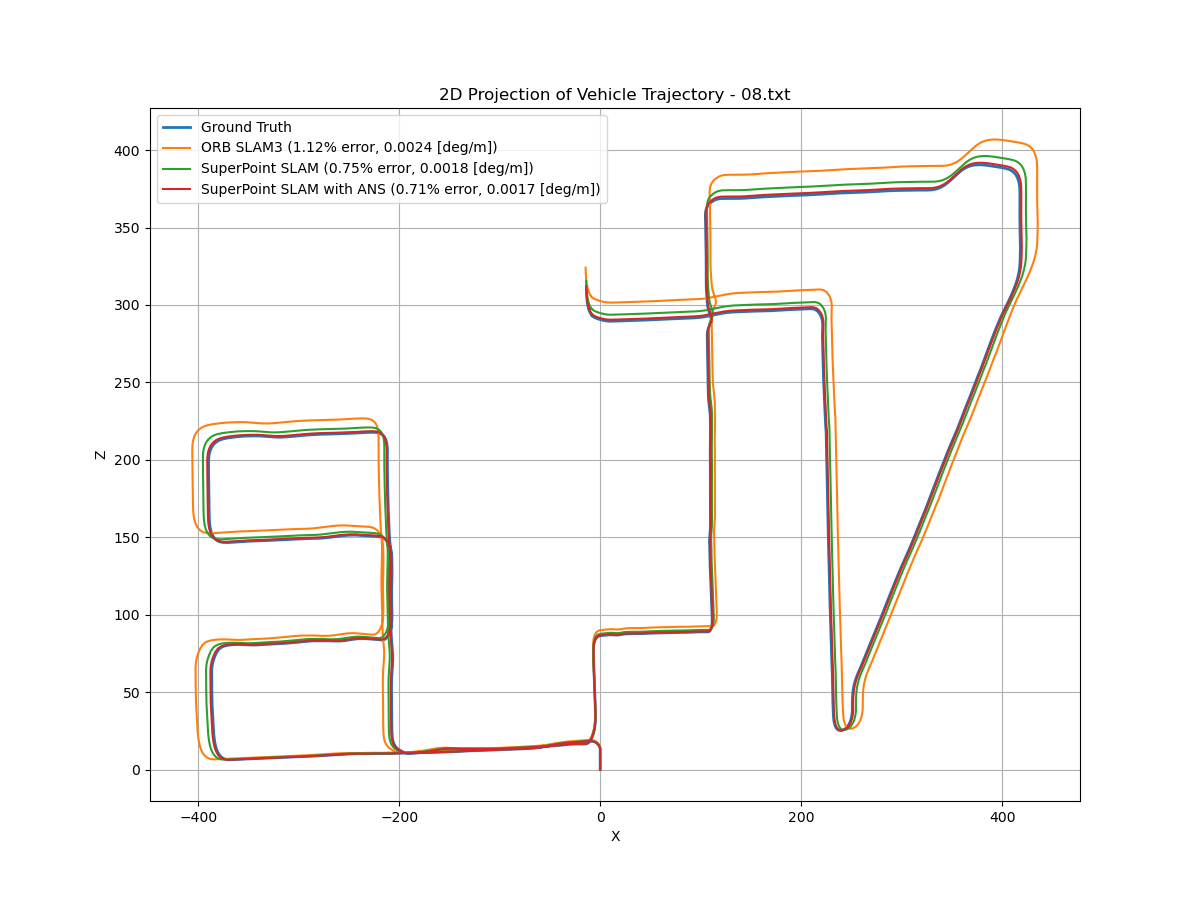} &
    \includegraphics[width=0.24\textwidth]{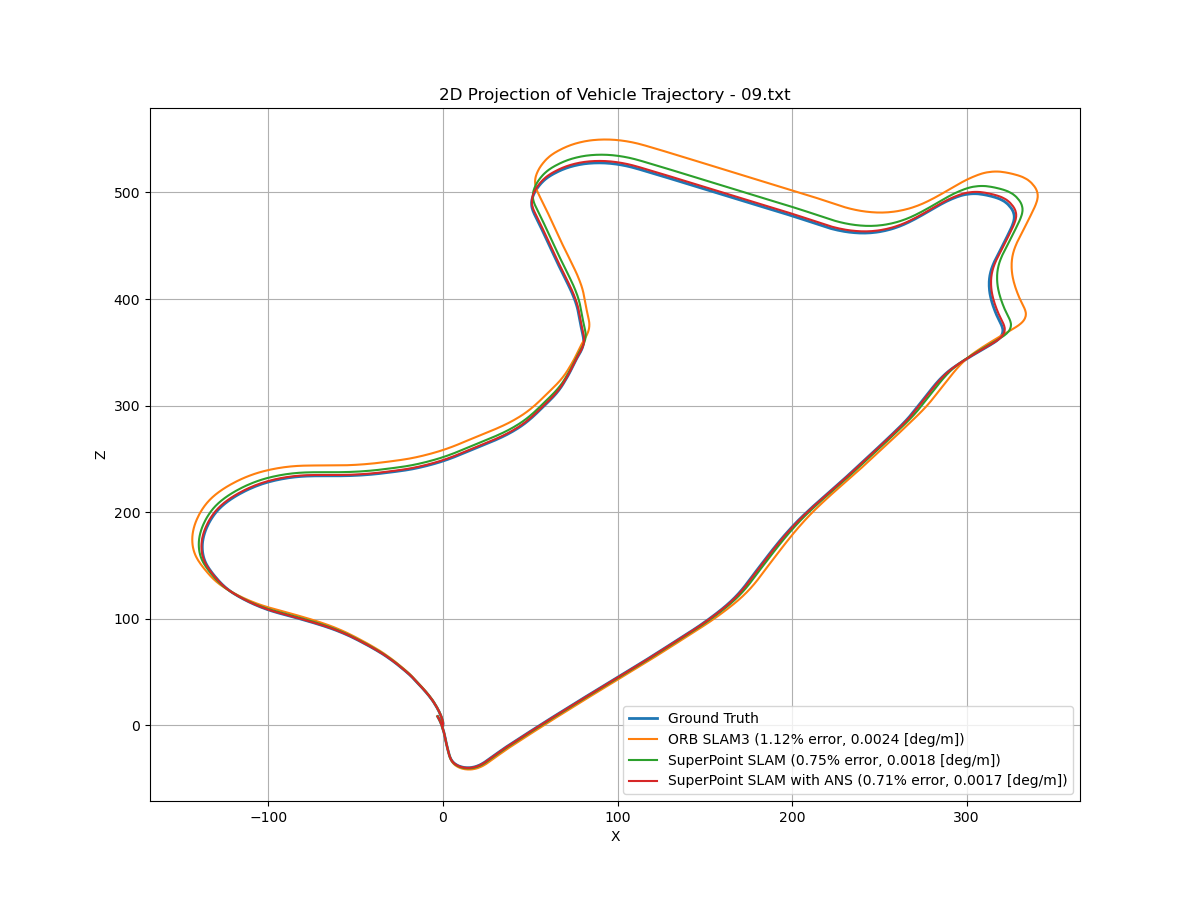} &
    \includegraphics[width=0.24\textwidth]{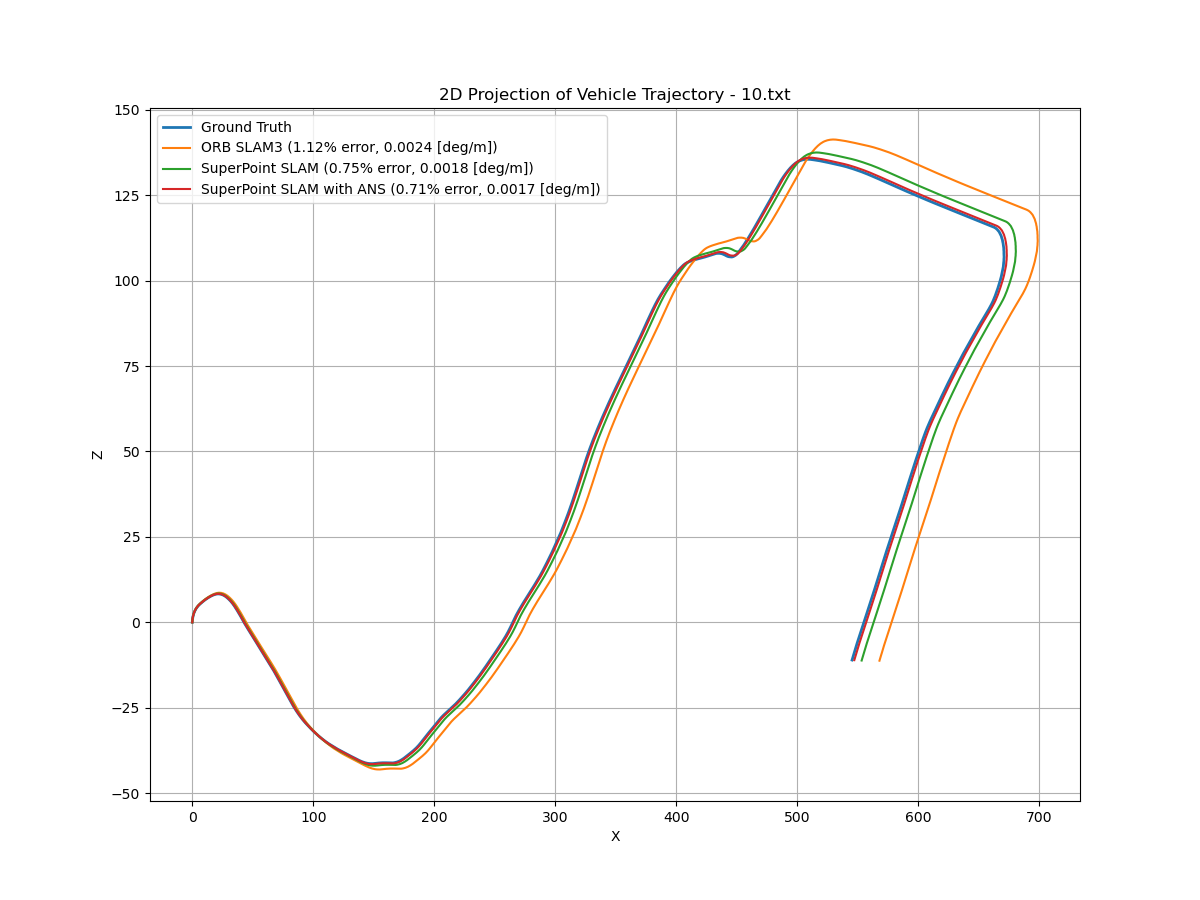} &
    \\
    Seq. 08 & Seq. 09 & Seq. 10 & \\
\end{tabular}
\caption{Comparative analysis of 2D trajectories (XZ plane) for sequences 00 to 10 using ORB-SLAM3, SuperPoint, and SuperPoint + ANMS.}
\label{fig:2d_traj}
\end{figure*}

The plots are shown in \ref{fig:2d_traj}



\subsection{6D Pose Estimation}

\begin{figure*}[ht]
\centering
\begin{tabular}{cccc}
    \includegraphics[width=0.24\textwidth]{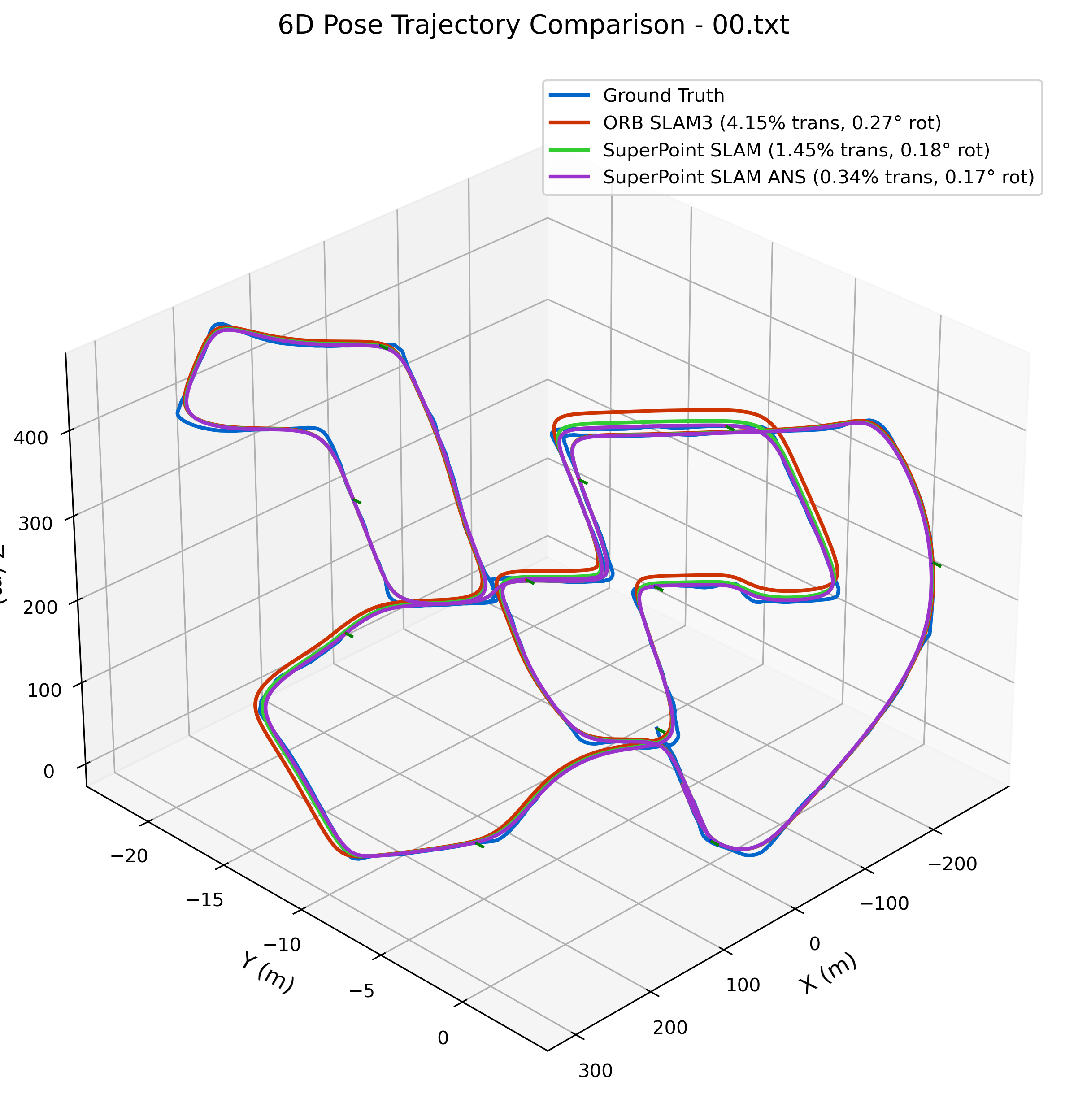} &
    \includegraphics[width=0.24\textwidth]{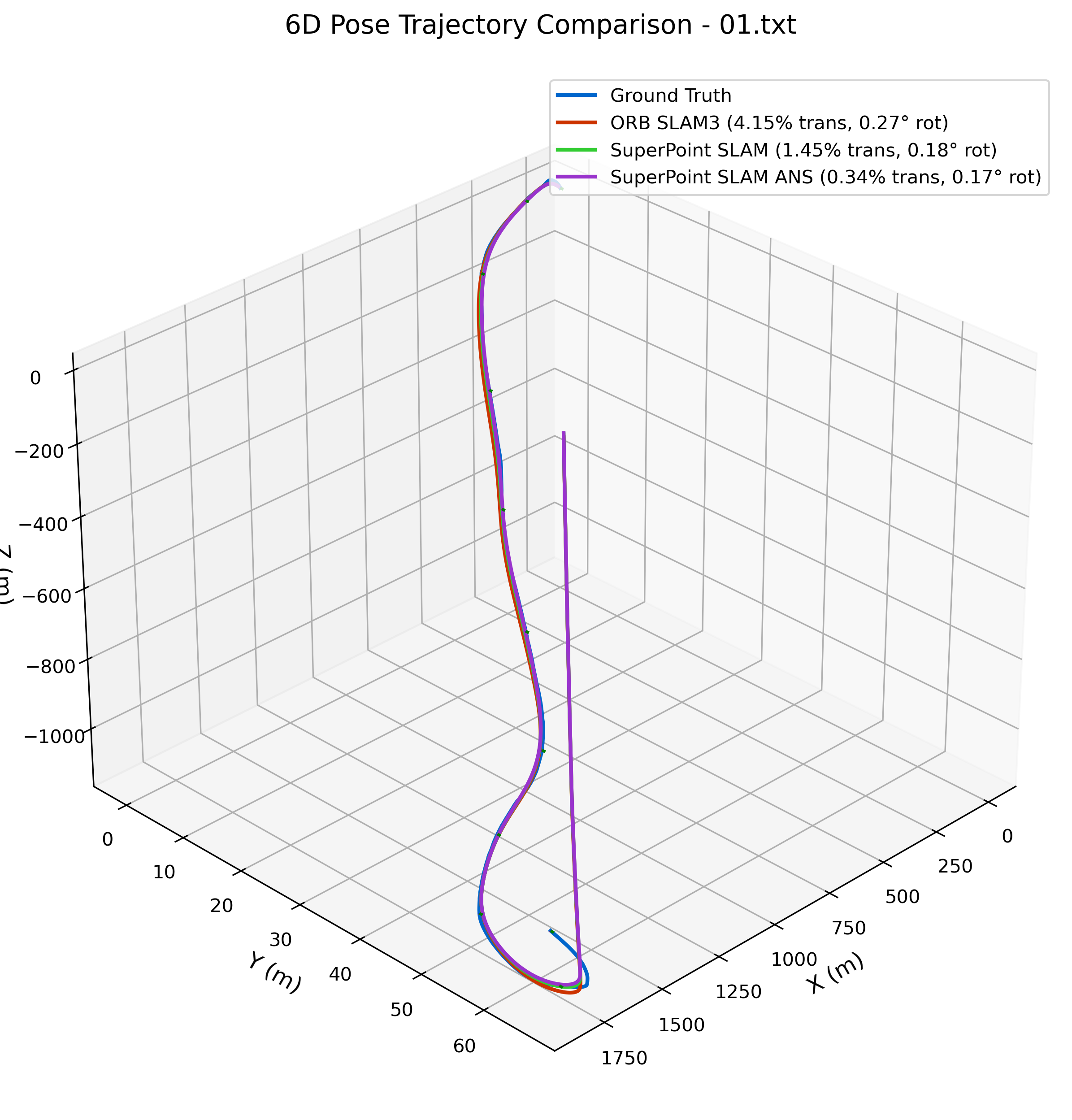} &
    \includegraphics[width=0.24\textwidth]{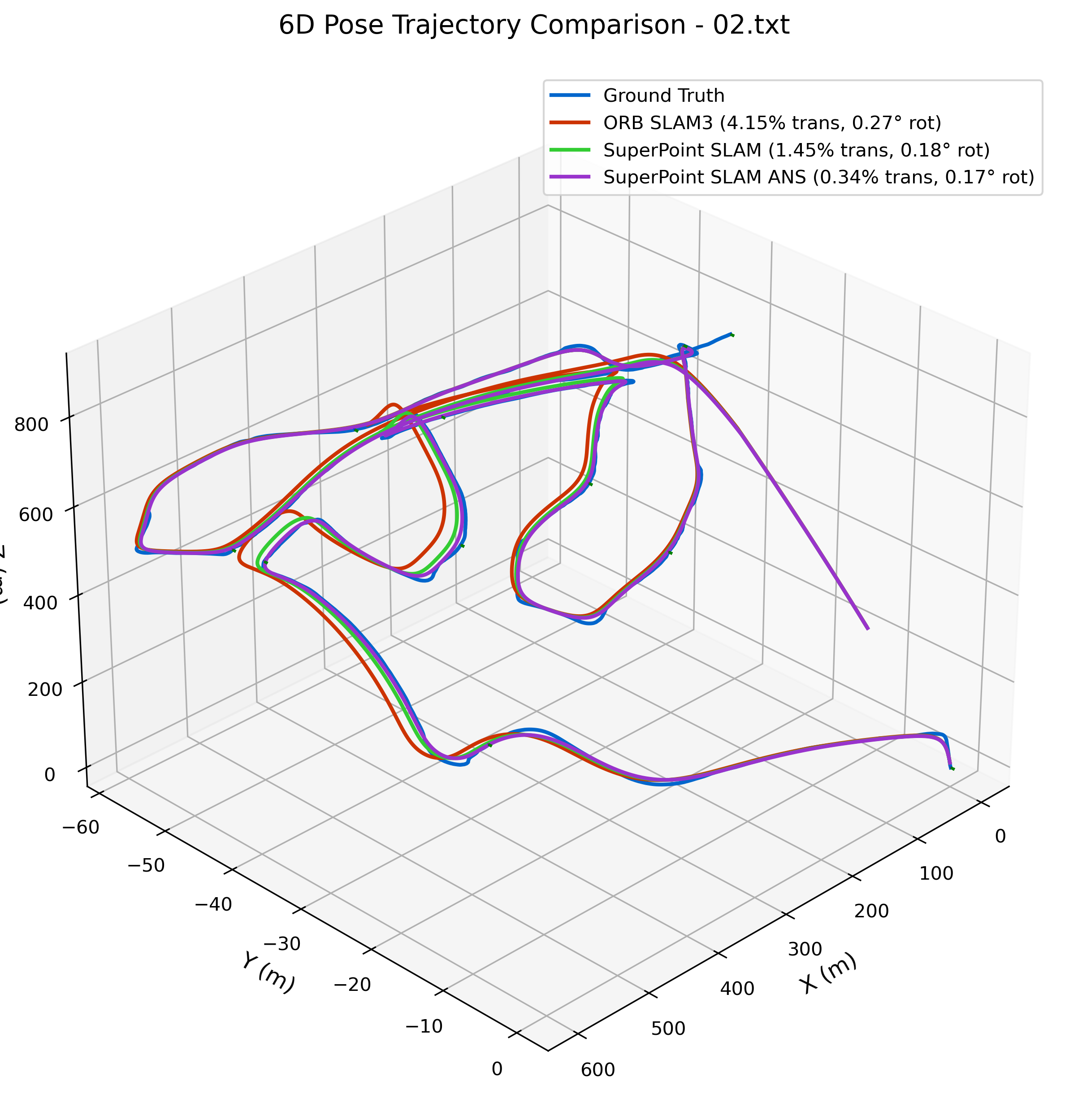} &
    \includegraphics[width=0.24\textwidth]{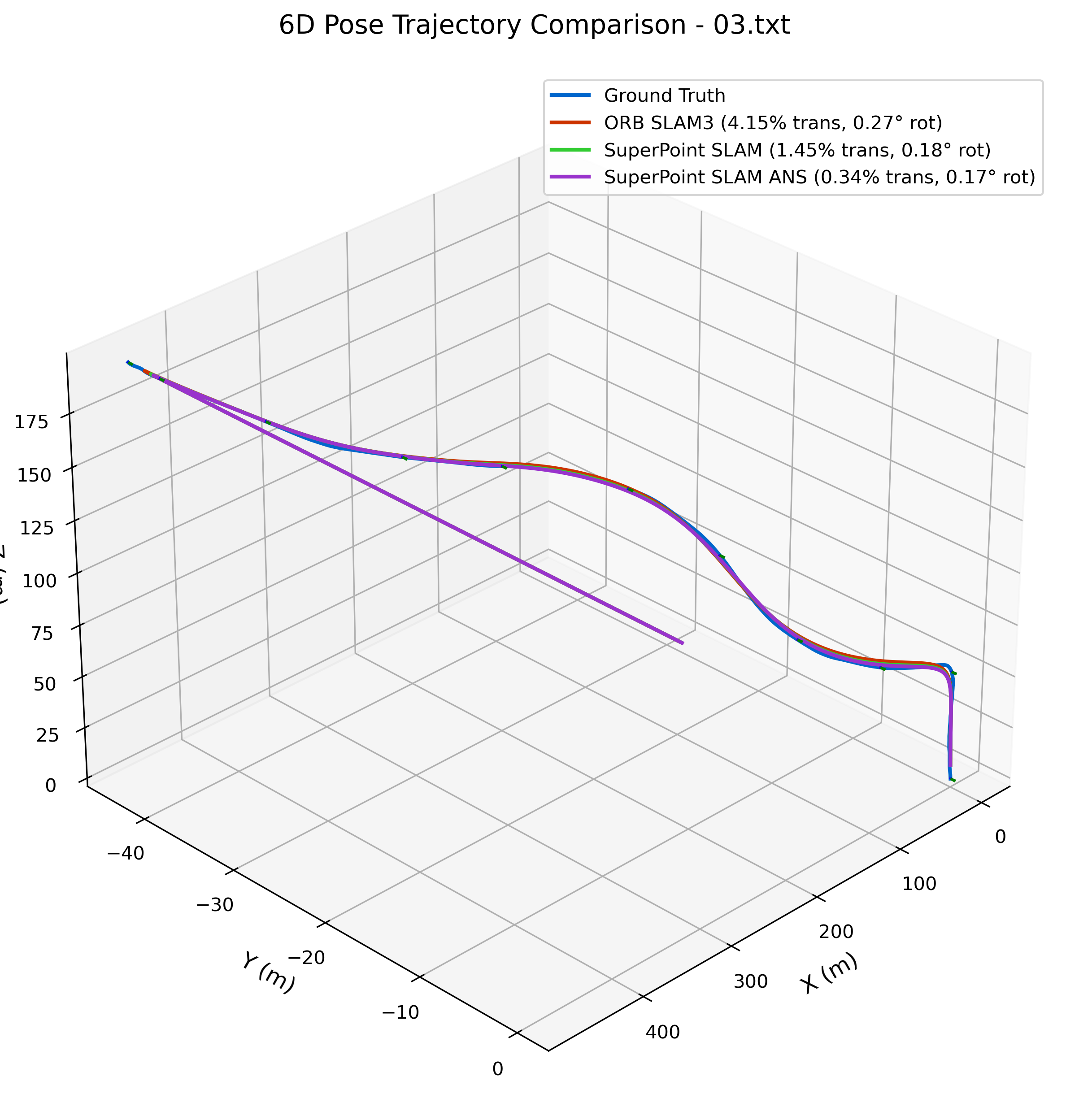} \\
    Seq. 00 & Seq. 01 & Seq. 02 & Seq. 03 \\[1em]
    \includegraphics[width=0.24\textwidth]{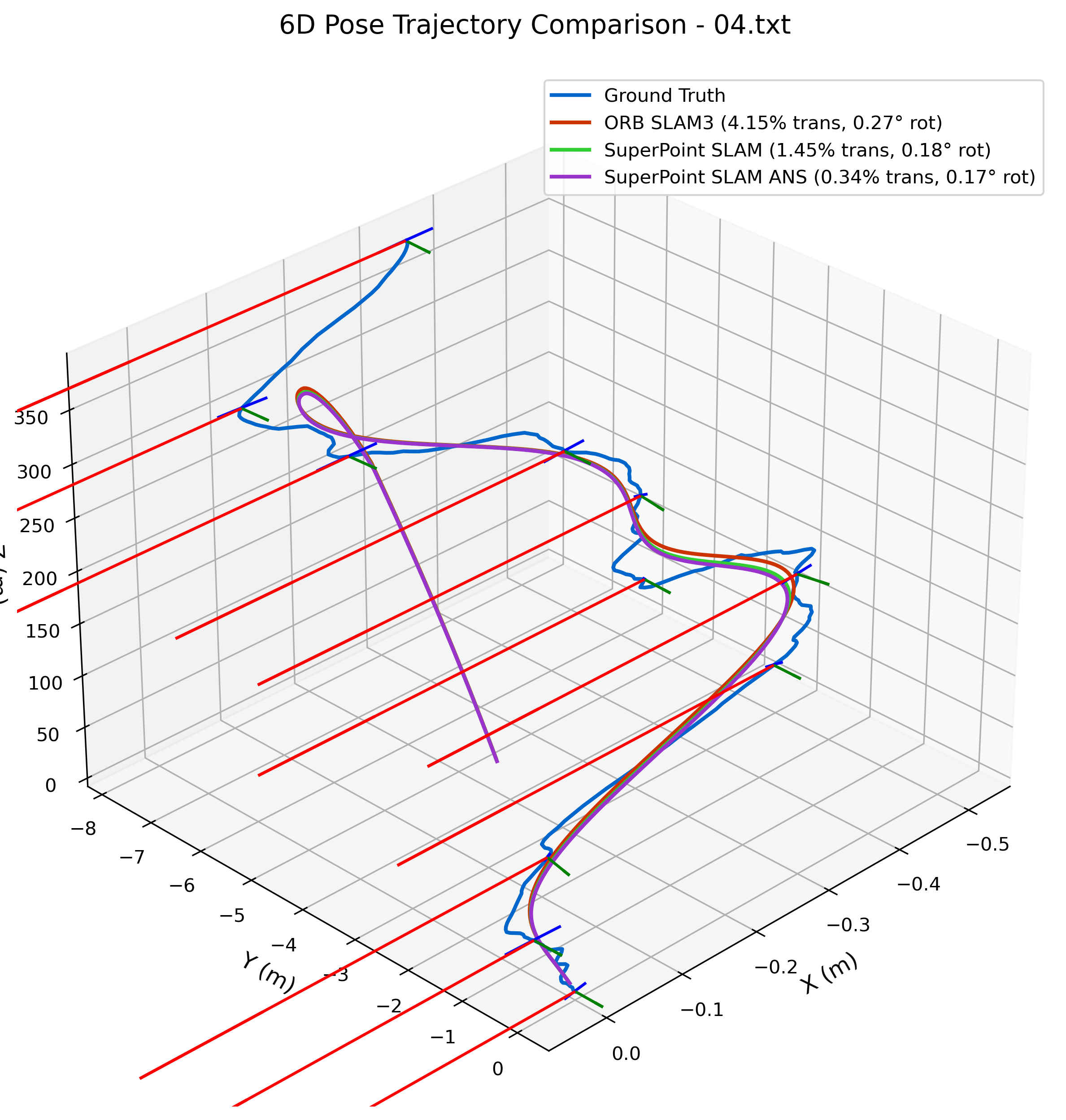} &
    \includegraphics[width=0.24\textwidth]{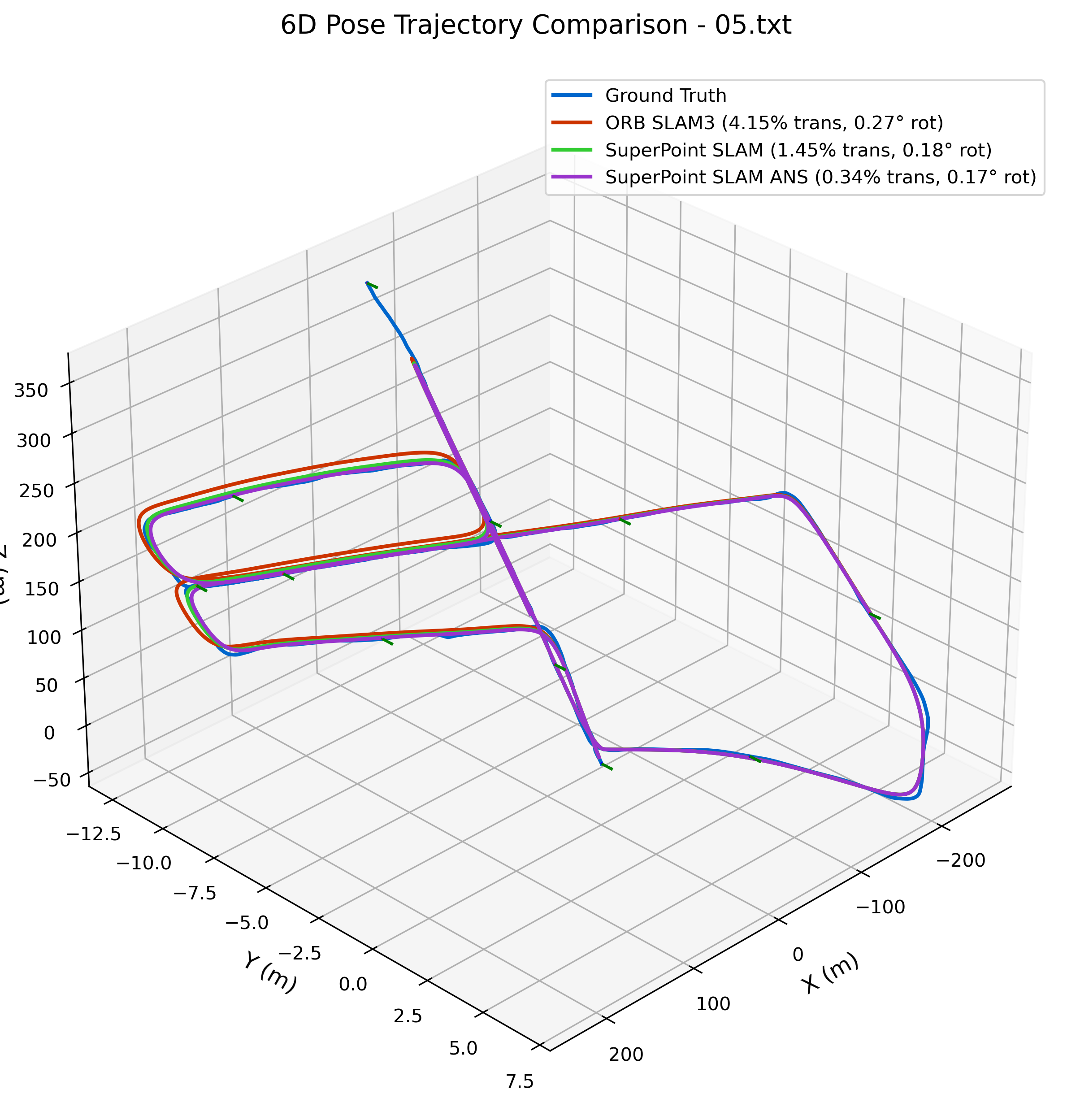} &
    \includegraphics[width=0.24\textwidth]{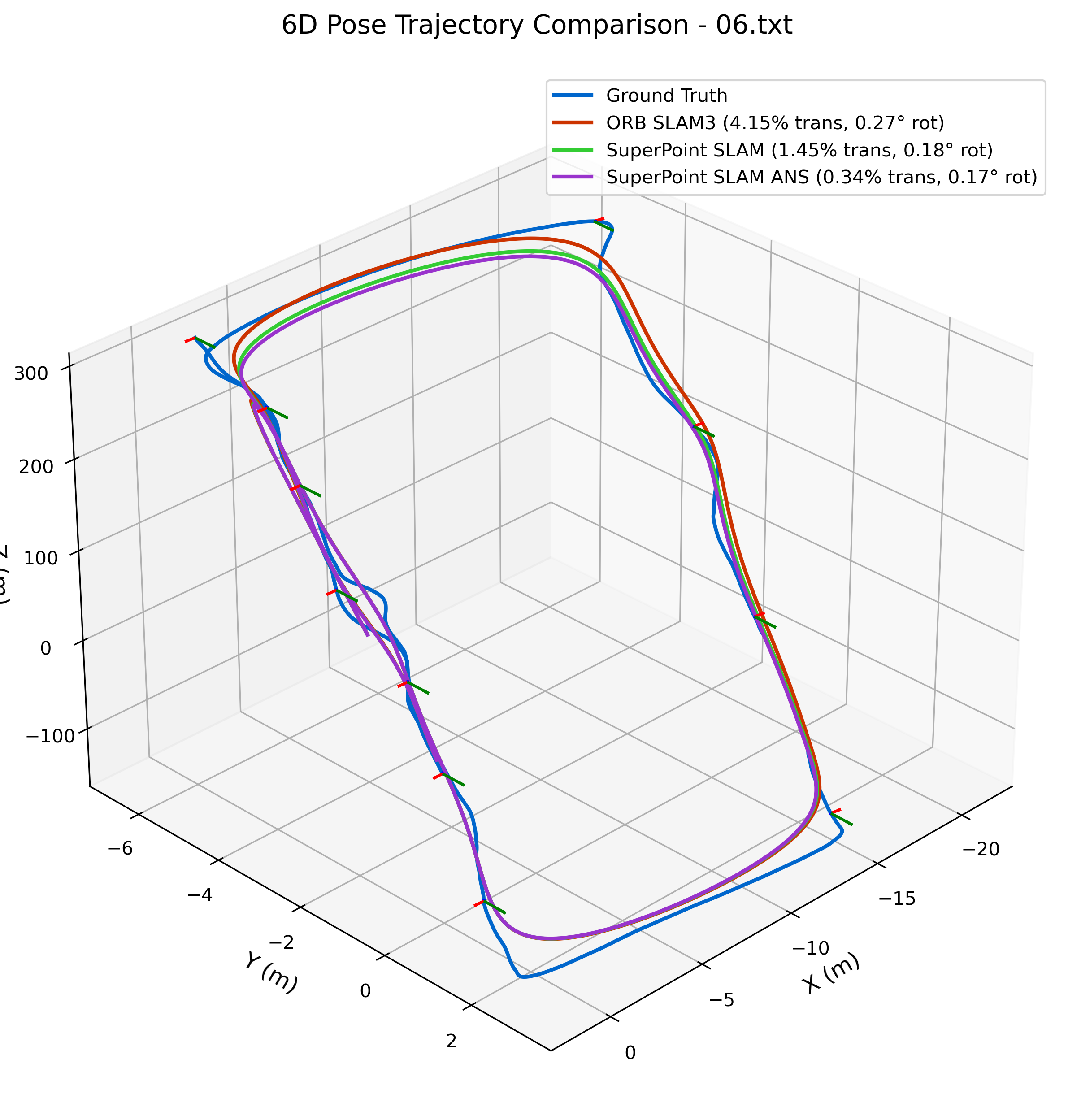} &
    \includegraphics[width=0.24\textwidth]{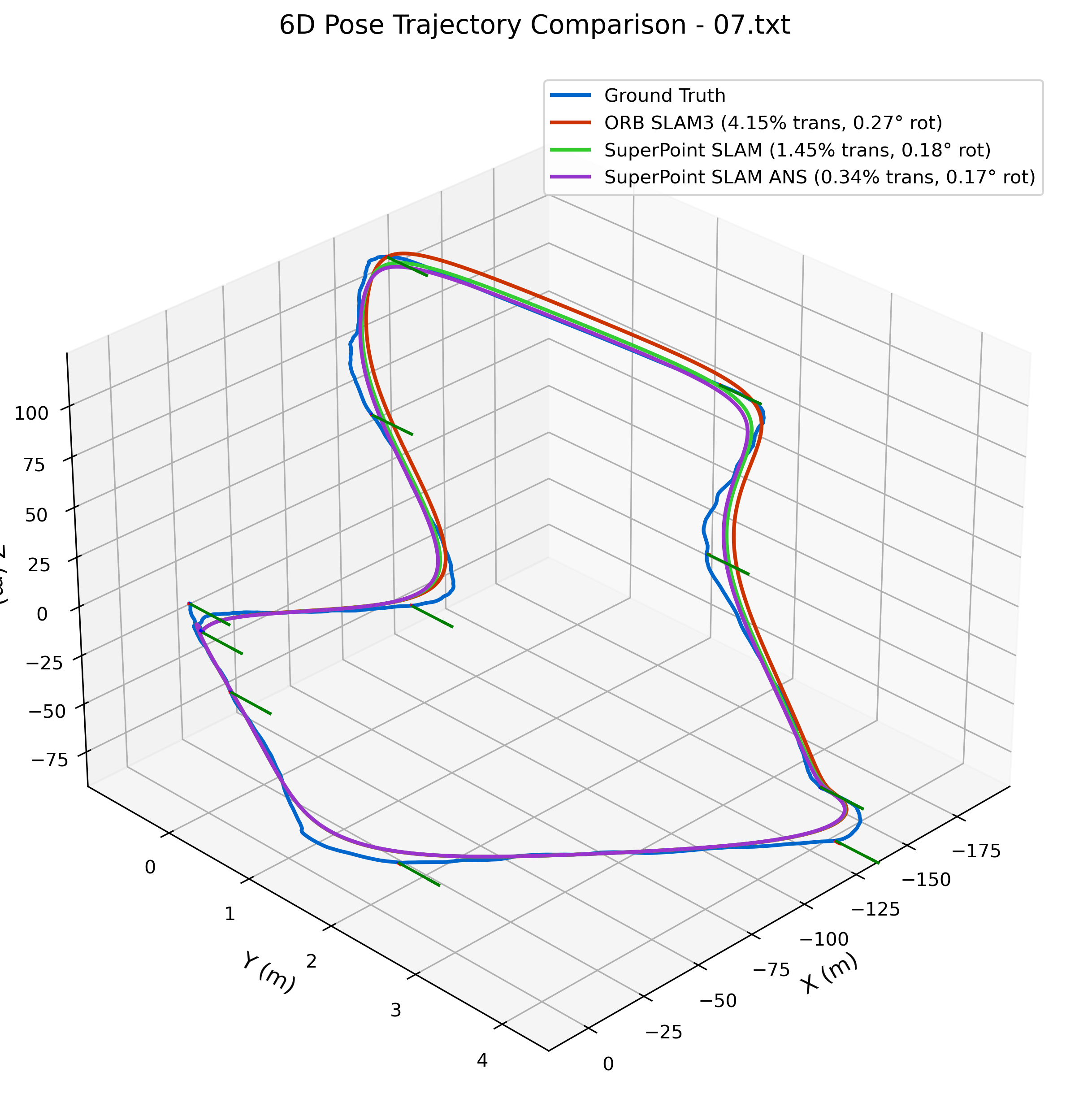} \\
    Seq. 04 & Seq. 05 & Seq. 06 & Seq. 07 \\[1em]
    \includegraphics[width=0.24\textwidth]{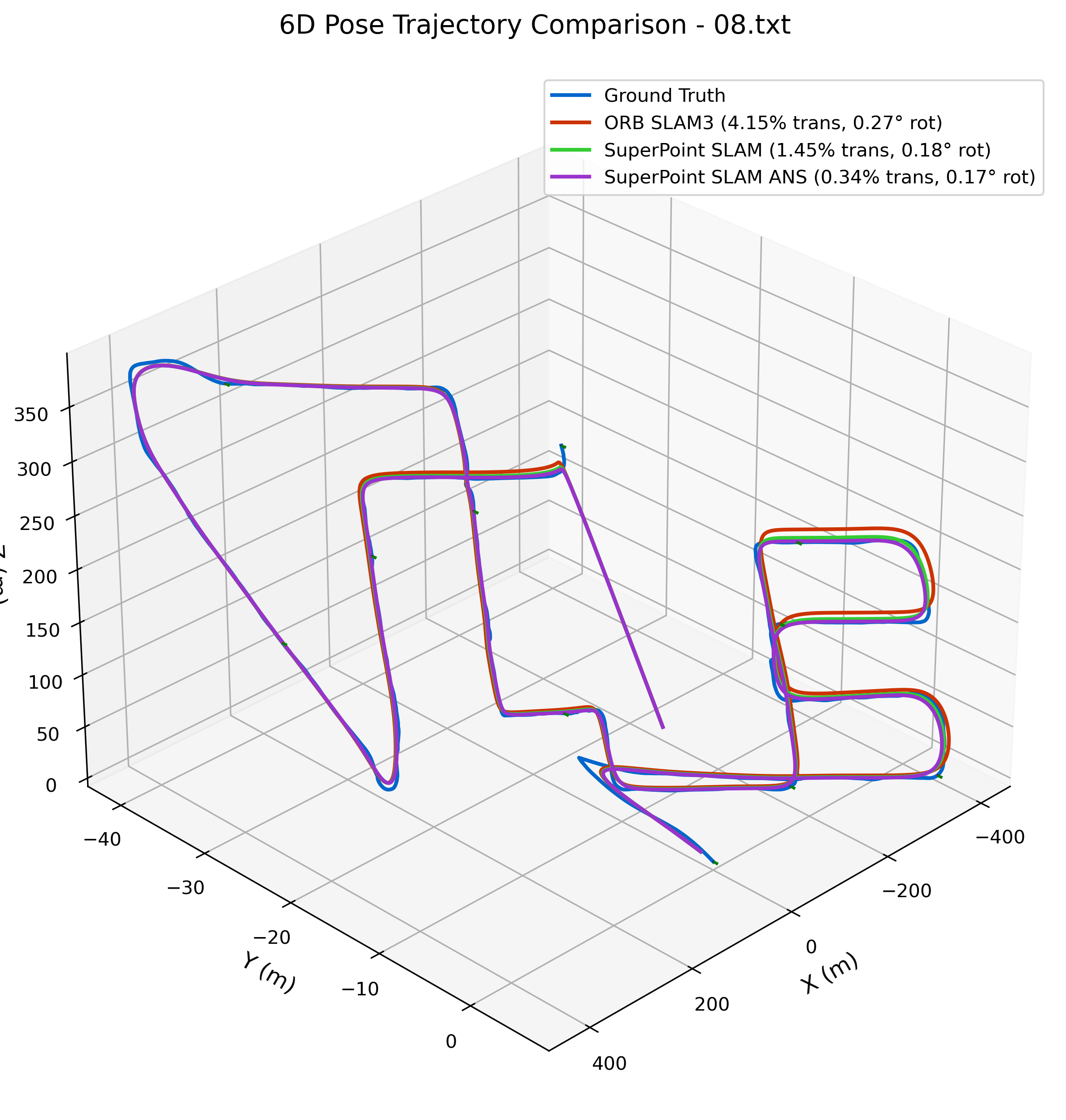} &
    \includegraphics[width=0.24\textwidth]{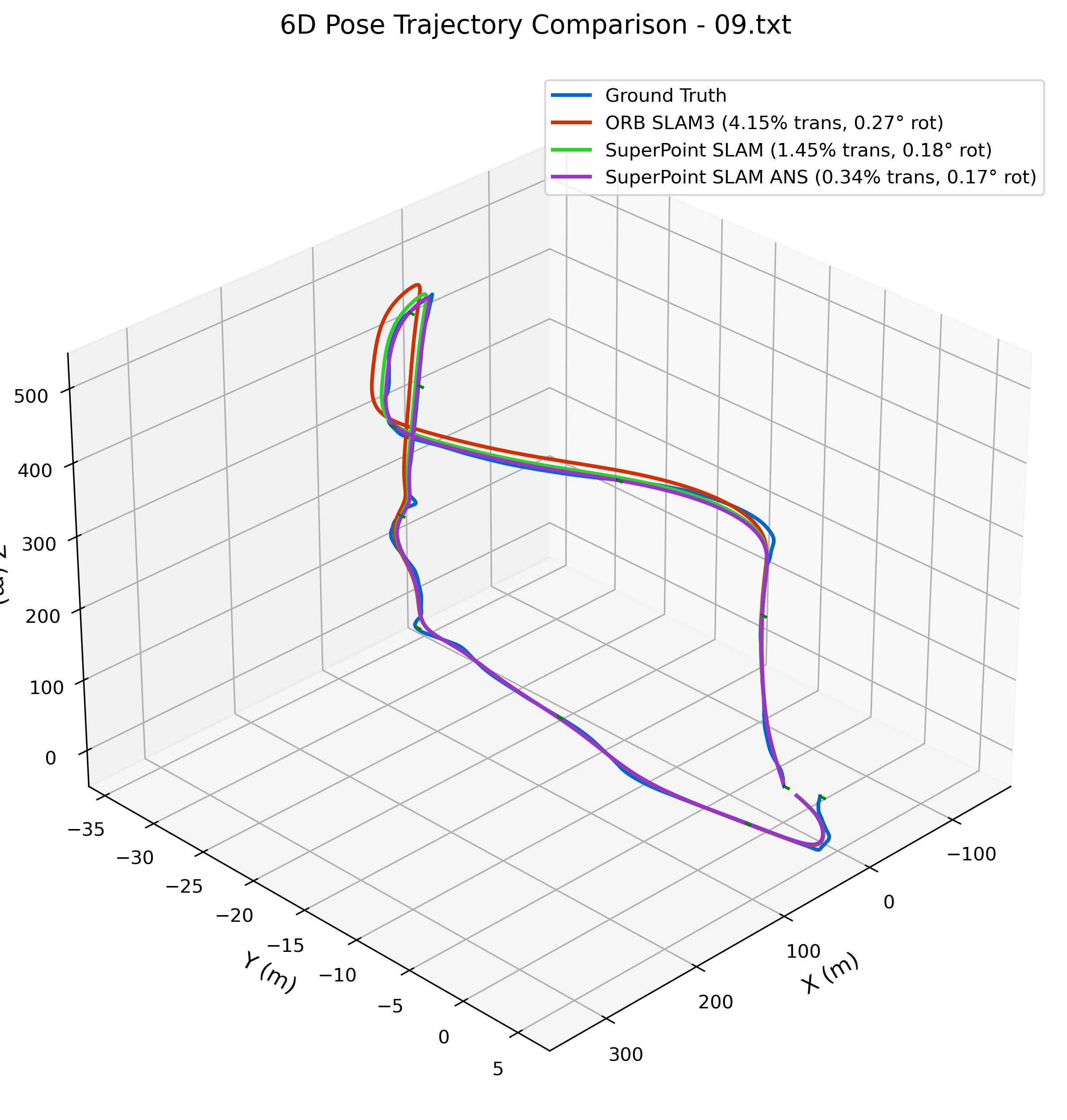} &
    \includegraphics[width=0.24\textwidth]{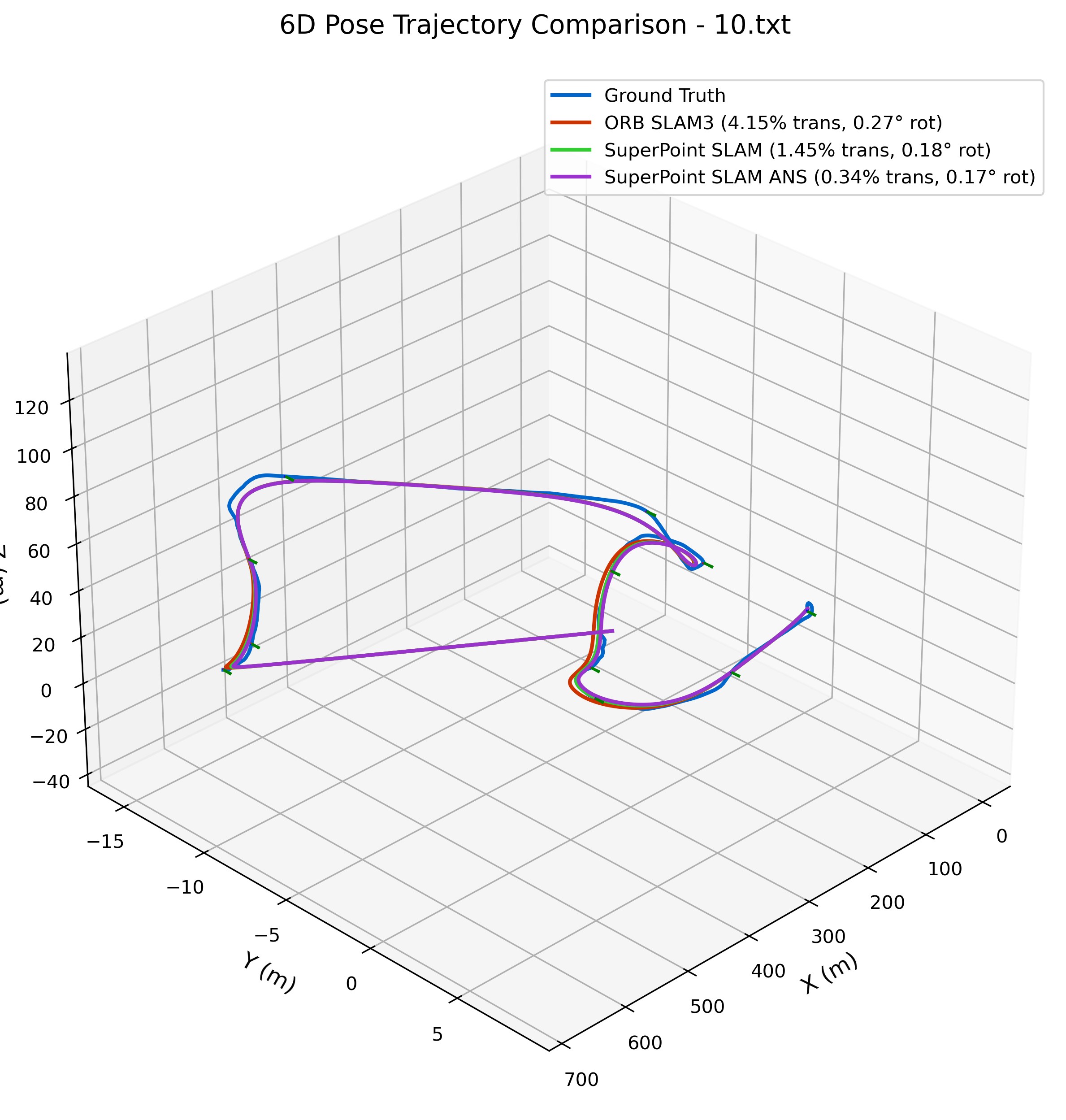} &
    \\
    Seq. 08 & Seq. 09 & Seq. 10 & \\
\end{tabular}
\caption{Comparative analysis of 6D pose estimation for ORB-SLAM3, SuperPoint, and SuperPoint + ANMS.}
\label{fig:6d_pose}
\end{figure*}

The plots are shown in \ref{fig:6d_pose}

\section{Limitations}

\subsection{Integration Complexity}
Modifying the ORB-SLAM3 codebase to integrate SuperPoint and ANMS was challenging due to differences in data structures and feature representations.

\subsection{Loop Closure Bug}
The system attempted incorrect loop closures in some sequences, indicating a need to adjust the loop closure detection mechanism for compatibility with SuperPoint features.

\subsection{Computational Load}
SuperPoint is more computationally intensive than ORB features, which may affect real-time performance. Optimization techniques and GPU acceleration may be required.

\section{Conclusion and Future Work}

Our modifications to ORB-SLAM3 by integrating SuperPoint features and ANMS have shown promising improvements in localization accuracy and robustness. The preliminary results suggest that deep learning-based features can enhance traditional SLAM systems.

\subsection{Future Work}
\begin{itemize}
    \item \textbf{Loop Closure Module:} Given that the SuperPoint feature descriptors are 256 bit float strings, and the BoWV2 operates on hamming distance, we need to address the loop closure bug by adjusting similarity metrics and thresholds by incorporating a learning-based loop closure mechanism.
    \item \textbf{Extended Evaluation:} Evaluate the system on additional datasets, including multi-modal datasets MS2 dataset, to assess performance in different scenarios and modalities.
\end{itemize}

\end{document}